\documentclass[10pt]{article}
\usepackage[letterpaper,margin=1.3in]{geometry} 
\usepackage{amsmath}
\usepackage{amssymb}
\usepackage{amsfonts}
\usepackage{booktabs}
\usepackage{tabularx}
\usepackage{amsmath}
\usepackage{enumitem}
\usepackage{array} 
\usepackage{graphicx}
\usepackage{subcaption} 
\usepackage{lipsum} 
\usepackage{pdfpages}
\usepackage{float} 
\usepackage{caption}
\usepackage{algorithm}
\usepackage{algpseudocode}

\usepackage[T1]{fontenc}
\usepackage{lmodern} 
\usepackage[hidelinks]{hyperref} 
\title{Neural operators struggle to learn complex PDEs in pedestrian mobility: Hughes model case study\\
}

\author{Prajwal Chauhan \thanks{NYUAD Research Institute, New York University Abu Dhabi, PO Box 129188, Abu Dhabi, United Arab Emirates, {\sf pc3377@nyu.edu}, {\sf sc8101@nyu.edu}, {\sf mg6888@nyu.edu}} \and Salah Eddine Choutri \footnotemark[1] \and Mohamed Ghattassi \footnotemark[1] \quad Nader Masmoudi \thanks{Division of Science, New York University in Abu Dhabi, Saadiyat Island, P.O. Box 129188, Abu Dhabi, United Arab Emirates, and Courant Institute of Mathematical Sciences, New York University, 251 Mercer Street, New York, NY 10012, USA, {\sf nm30@nyu.edu}}\and Saif Eddin Jabari \thanks{Division of Engineering, New York University Abu Dhabi, PO Box 129188, Abu Dhabi, United Arab Emirates, and New York University Tandon School of Engineering, Brooklyn, New York 11201, USA, {\sf sej7@nyu.edu}}}
\date{}
\begin{document}
\maketitle
\begin{abstract}
This paper investigates the limitations of neural operators in learning solutions for a Hughes model, a first-order hyperbolic conservation law system for crowd dynamics. The model couples a Fokker–Planck equation, representing pedestrian density, with a  Hamilton–Jacobi type equation (the eikonal equation). This Hughes model belongs to the class of nonlinear hyperbolic systems which often exhibit complex solution structures, including shocks and discontinuities. In this study, we assess the performance of three state of the art neural operators Fourier Neural Operator (FNO), Wavelet Neural Operator (WNO), and Multiwavelet Neural Operator (MWT) in various challenging scenarios. Specifically, we consider (1) discontinuous and Gaussian initial conditions and (2) diverse boundary conditions while also examining the impact of different numerical schemes.
Our results show that these neural operators perform well in easy scenarios characterized by fewer discontinuities in the initial condition, yet, they struggle in complex scenarios with multiple initial discontinuities and dynamic boundary condition, even when trained specifically on such complex samples.
The predicted solutions often appear smoother, resulting in a reduction in total variation and a loss of important physical features. This smoothing behavior is similar to issues discussed in \cite{daganzo1995requiem}, where models that introduce artificial diffusion were shown to miss essential features such as shock waves in hyperbolic systems. This suggests that current neural operator architectures may introduce unintended regularization effects that limit their ability to capture transport dynamics governed by discontinuities. Since the Hughes model shares important structural features with models used in traffic flow, these results also raise concerns about the ability of neural operator architectures to generalize to traffic applications where shock preservation is essential.

\end{abstract}

 \section{Introduction}
The remarkable progress in deep learning and artificial intelligence has significantly transformed computational approaches across various scientific and engineering disciplines, particularly in the context of solving partial differential equations (PDEs) \cite{raissi2019physics, lu2021learning}. Traditional numerical methods, while robust and widely used, often face substantial challenges when dealing with nonlinear hyperbolic PDEs. These PDEs are characterized by complex features such as shocks, discontinuities, and steep gradients \cite{leveque2002finite}, which can give rise to numerical instabilities, increased computational costs, and difficulties in achieving accurate solutions. Such limitations motivate the exploration of alternative methodologies that leverage the capabilities of machine learning. One promising direction involves neural operator frameworks, which have recently shown the ability to approximate PDE solutions in a mesh-independent manner and to generalize across different instances of specific PDE problems \cite{kovachki2021neural, li2020fourier, jabari2022learning, thodi2024fourier}. However, despite these advances, neural operators remain relatively unexplored in the context of nonlinear inviscid hyperbolic PDEs, particularly in realistic scenarios featuring intricate boundary conditions and solution discontinuities.

Recent work in transportation has highlighted the potential of neural operators to model complex dynamics. They have been applied to traffic flow problems \cite{morand2024deep} and traffic state estimation tasks \cite{gao2024comparative, huang2020physics}. Nevertheless, challenges remain in preserving essential features of hyperbolic systems. In \cite{daganzo1995requiem}, artificial diffusion in continuum traffic models was shown to suppress key structures such as shock waves. These observations raise concerns about the ability of current neural operator architectures to generalize in systems that require accurately preserving discontinuities.
In this paper, we explore how neural operators can be applied to the Hughes model, a nonlinear hyperbolic PDE system describing pedestrian dynamics through two coupled equations: a conservation law for pedestrian density and an eikonal equation governing pedestrian velocity. Owing to these coupled equations, the model naturally exhibits shocks and sharp discontinuities \cite{hughes2002continuum}. Originally proposed by Hughes \cite{hughes2002continuum} to represent pedestrian movement and evacuation in a two-dimensional space, the model has been the subject of extensive research, we refer to the survey on the numerical and analysis of the hughes model \cite{amadori2023mathematical}. More recently, Ghattassi and N. Masmoudi \cite{gm2023} expanded the Hughes framework by incorporating mean-field games (MFGs) to capture pedestrian behavior that avoids high-density regions; they also investigated the stability of this generalized model \cite{ghattassi2023non}. Next, we introduce the mathematical Hughes model of the macroscopic pedestrian flow. The pedestrian density \(\rho=\rho(t,x)\) evolves according to a scalar conservation law, and the preferred walking direction is determined by an eikonal equation.  
Specifically, on the interval $\Omega = (-1,1)$, the model is given by
\begin{subequations}\label{eq:1.1}
\begin{align}
\label{eq:1.1a}
& \rho_t - \Bigl(\rho\,v(\rho)\,\frac{\phi_x}{|\phi_x|}\Bigr)_x = 0,\\[6pt]
\label{eq:1.1b}
& |\phi_x| = c(\rho),
\end{align}
\end{subequations}
where $x \in \Omega$ is the spatial coordinate, $t \ge 0$ represents time, and $\rho(t,x)$ ranges in $[0,1]$ (interpreted as a normalized density).  
The cost function $c(\rho)$ is assumed to be smooth from $[0,1]$ into $[1,+\infty[$, with $c(0)=1$ and $c'(\rho)\ge 0$.  
We also set the mean velocity as $v(\rho) = (1-\rho)$ and define
\[
f(\rho) = \rho\,v(\rho) = \rho\,(1-\rho).
\]
In many studies (see \cite{di2011hughes}), a commonly adopted choice is
\begin{equation}
\label{eq:1.2}
c(\rho) = \frac{1}{v(\rho)} 
= \frac{1}{1-\rho}.
\end{equation}

To complete the model, we prescribe an initial condition \(\rho(0,\cdot) = \rho_0 \in BV(\mathbb{R})\) and non homogeneous Dirichlet boundary data at $x = \pm 1$.
Concretely, we set
\begin{subequations}\label{eq:1.3}
\begin{align}
\label{eq:1.3a}
& \rho(t,-1) =\rho_{bc1}(t), \qquad  \rho(t,1) = \rho_{bc2}(t),
\\[6pt]
\label{eq:1.3b}
& \phi(t,-1) =\phi_{bc1}(t), \qquad    \phi(t,1) = \phi_{bc2}(t).
\end{align}
\end{subequations}
For a discussion of more complex boundary conditions, we refer the reader to \cite{colombo2011modelling, goatin2006awavefront, amadori2023mathematical} and the references therein. Noting that \eqref{eq:1.1a} can be rearranged as
\begin{equation}
\label{eq:1.4}
\rho_t - \Bigl(f(\rho)\,\mathrm{sgn}(\phi_x)\Bigr)_x = 0,
\end{equation}
one finds that the unique viscosity solution of the Dirichlet problem \eqref{eq:1.1b}, \eqref{eq:1.3b} can be characterized by the value function of a control problem with the (possibly) discontinuous coefficient \(c(\rho)\).  
In particular,
\begin{equation}
\label{eq:1.5}
\phi(t,x)
=
\begin{cases}
\displaystyle 
\int_{-1}^{x} c\bigl(\rho(t,y)\bigr)\,dy + \phi_{bc1}(t),
& \text{if } -1 \le x \le \xi(t),
\\[6pt]
\displaystyle
\int_{x}^{\,1} c\bigl(\rho(t,y)\bigr)\,dy + \phi_{bc2}(t),
& \text{if } \xi(t) \le x \le 1,
\end{cases}
\end{equation}
where \(\xi(t)\in \Omega\) is implicitly determined via

\begin{equation} \label{eq:hughes_mod}
\rho_t + \partial_x \left[\text{sgn}(x - \xi(t)) \rho (1 - \rho) \right] = 0,
\end{equation}
where $\xi(t)$ is the turning point, which can be determined by the cost integral given by
\begin{equation} \label{eq:cost_integral}
\int_{-1}^{\xi(t)} c\bigl(\rho(t,y)\bigr)+\phi_{bc1}(t)\ dy 
=
\int_{\xi(t)}^{1} c\bigl(\rho(t,y)\bigr)+\phi_{bc2}(t)\ dy.
\end{equation}
Obtaining smooth solutions for the Hughes model is numerically challenging due to the formation of shocks and discontinuities in crowd dynamics. These complexities necessitate robust numerical methods that can handle abrupt changes in the solution. In this work, we generate training datasets by numerically solving Eq. \eqref{eq:hughes_mod} along with Eq. \eqref{eq:cost_integral} which balances the cumulative cost of movement toward preferred direction. The turning point, here, $\xi(t)$ represent the location at which the cost is same to move toward any direction. If the pedestrians at time $t$ happen to be in a location that is left of $\xi(t)$, they exit left and vice versa.

\subsection{Neural operators: Background}
 Neural operators have emerged as a promising class of machine learning frameworks for approximating solution operators to partial differential equations (PDE's). One of the earliest operator learning frameworks is DeepONet Lu et al. \cite{lu2021learning} which uses two subnetworks, a branch and a trunk network, to approximate nonlinear operators and has inspired various extensions. Following this direction, Li et al. \cite{kovachki2021neural} proposed the neural operator framework. One of their early models was an iterative architecture inspired by Green’s functions of elliptic PDEs involving an integral kernel which is computed by message passing on graph networks\cite{li2020neuralb}. The same authors later extended this with a multipole-based design over multiscale graph structures, capturing interaction at all ranges with only linear complexity \cite{li2020multipole}.

The Fourier Neural Operator (FNO) \cite{li2020fourier} marked a significant advance by learning mappings in Fourier space to capture global dependencies. It has been tested on Darcy flow (elliptic PDE with Dirichlet boundary conditions), the 2D Navier--Stokes equations with periodic boundary conditions, and the 1D viscous Burgers equation. These settings use smooth initial conditions and periodic geometries, avoiding the challenges of discontinuous, inviscid, or highly nonlinear regimes.

The Multiwavelet Neural Operator (MWT) \cite{gupta2021multiwavelet} extended the idea using orthonormal multiwavelet bases, offering better localization in space and frequency. MWT was applied to the Korteweg-de Vries (KdV) equation, Burgers equation, Darcy flow, and 2D Navier--Stokes equations, generally under periodic or smoothed settings, again simplifying the solution landscape.

The Wavelet Neural Operator (WNO) \cite{tripura2022wavelet} introduced wavelet transforms to improve localization. Tested on Darcy flow, the viscous and inviscid Burgers equations, wave advection, and the Allen--Cahn equation, WNO showed improved resolution of localized features. Yet, the configurations often involved low-amplitude or smoothed inputs, and structured grids with periodic boundaries.

The Pseudo-Differential Neural Operator (PDNO) \cite{shin2024pseudodifferential} generalizes the spectral kernel of FNO by parameterizing pseudo-differential operators, improving adaptability to spatial heterogeneity. Still, PDNO has been tested mostly on smooth Darcy  flow and Navier-Stokes equation with "nice" set of paramters.

Koopman Neural Operator (KNO) \cite{xiong2024koopman} maps nonlinear PDE dynamics to a latent linear system using Koopman theory. It has been tested on relatively simple systems like reaction-diffusion equations with smooth dynamics.

While many neural operators have demonstrated strong performance in approximating PDE solutions under smooth, periodic, or low-parameter regimes, their application to more realistic nonlinear hyperbolic PDEs with discontinuous solutions, non-periodic boundaries, or complex geometries remains limited. The use of periodic boundaries, moderate Reynolds numbers, and smoothed initial data in most benchmarks masks many of the challenges that would arise in more difficult, real-world applications. This paper investigates such limitations by testing neural operators on the Hughes model, a challenging nonlinear hyperbolic system for pedestrian dynamics where sharp fronts and solution discontinuities naturally occur.

\subsection{Neural operators: Overview}
Neural operators provide a framework for learning mappings between infinite-dimensional function spaces, making them well-suited for learning solutions to certain parametric partial differential equations (PDEs). They can be  advantageous in scientific and engineering contexts where repeated PDE evaluations are required. Foundational works such as \cite{li2020fourier, kovachki2021neural} have established the theoretical and architectural underpinnings of neural operators.

Their goal is to approximate a nonlinear operator \( G^\dagger: A \to U \), where \( A = A(D; \mathbb{R}^{d_a}) \) and \( U = U(D; \mathbb{R}^{d_u}) \) are Banach spaces of functions defined over a bounded domain \( D \subset \mathbb{R}^d \). Given a dataset of input-output pairs \( \{a_j, u_j\}_{j=1}^N \), with \( a_j \sim \mu \) and \( u_j = G^\dagger(a_j) \), the goal is to learn a parametric approximation \( G_\theta: A \to U \), where \( \theta \in \Theta \) belongs to a finite-dimensional parameter space. This is achieved by minimizing the expected discrepancy between predictions and ground truth:
\[
\min_{\theta \in \Theta} \mathbb{E}_{a \sim \mu} \big[C(G(a, \theta), G^\dagger(a))\big].
\]

Neural operators are typically implemented using iterative update architectures. The input function \( a \in A \) is first mapped to a feature space via a local transformation \( v_0(x) = P(a(x)) \), with \( P \) often realized as a fully connected neural network. Then, for each iteration \( t = 0, \dots, T-1 \), the representation is updated according to:
\begin{equation}
    v_{t+1}(x) := \sigma\big(Wv_t(x) + (K(a; \phi)v_t)(x)\big), \quad \forall x \in D,
    \label{NO-Layer}
\end{equation}
where \( W: \mathbb{R}^{d_v} \to \mathbb{R}^{d_v} \) is a learned linear operator, \( \sigma \) is a non-linear activation function, and \( K(a; \phi) \) is a kernel integral operator parameterized by \( \phi \). The final output \( v_T(x) \) is projected back to the target space \( U \) through another local transformation \( Q \).

The kernel integral operator \( K(a; \phi) \) captures non-local dependencies and is defined as:
\begin{equation} \label{Kernel}
(K(a; \phi)v_t)(x) := \int_D \kappa(x, y, a(x), a(y); \phi)v_t(y) \, dy, \quad \forall x \in D,
\end{equation}
where \( \kappa: \mathbb{R}^{2(d + d_a)} \to \mathbb{R}^{d_v \times d_v} \) is a learnable kernel function. This construction extends classical neural network operations to infinite-dimensional spaces via compositions of integral operators and pointwise nonlinearities.

\begin{figure}
        \centering
        \includegraphics[width=0.95\linewidth]{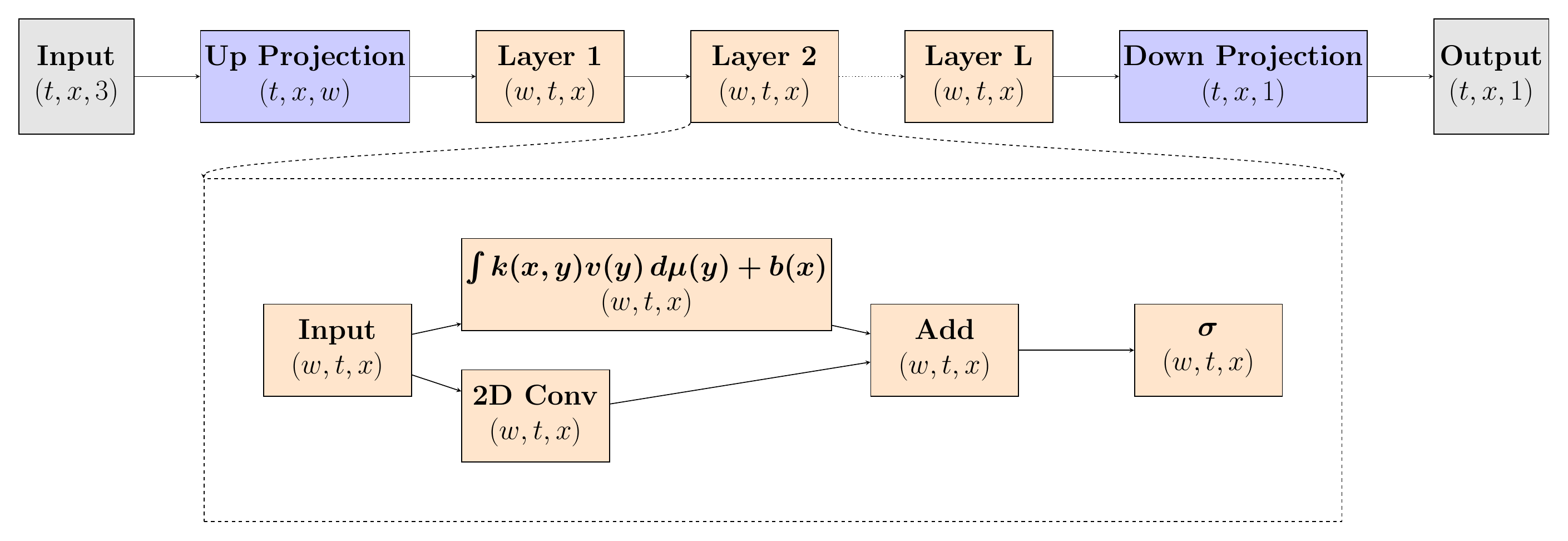}
        \caption{Layer-wise architecture of Neural Operator.}
        \label{fig:archi}
    \end{figure}

Each neural operator computes \eqref{Kernel} differently, trying to achieve a desired accuracy while reducing computational time. For instance, the Fourier neural operator imposes the restriction \( \kappa(x, y, a(x), a(y); \phi) = \kappa(x - y; \phi) \), converting the kernel integral operator into a convolution operator. This structure allows fast computation using the Fast Fourier Transform (FFT) (see figure~\ref{fig:archi}). Indeed, by the convolution theorem, we have:
\begin{equation*}
    (f * g)(x) = \mathcal{F}^{-1}(\mathcal{F}(f) \cdot \mathcal{F}(g))(x).
\end{equation*}

Figure~\ref{fig:archi} illustrates the data processing pipeline of a neural operator (NO). The NO accepts input with shape \((t,x,3)\), meaning that for each \((t,x)\) pair, there are three values corresponding to the input variable, the time coordinate, and the space coordinate. This input is first projected into a higher-dimensional space, increasing the feature dimension from 3 to a width \(w\) (with \(w > 3\)). The resulting representation is then permuted to the shape \((w,t,x)\) and then processed through multiple operator layers. The output of the last operator layer \(L\) is then permuted again to \((t,x,w)\) and finally down-projected to produce an output with shape \((t,x,1)\), so that each \((t,x)\) pair corresponds to a single output value.

To systematically examine the capabilities and limitations of neural operators, we perform extensive numerical experiments using three state-of-the-art neural operator architectures: Fourier Neural Operator (FNO) \cite{li2020fourier}, Wavelet Neural Operator (WNO)\cite{tripura2022wavelet}, and Multiwavelet Neural Operator (MWNO) \cite{gupta2021multiwavelet}. Our evaluation encompasses various  scenarios, including multiple discontinuous initial conditions, different boundary conditions and two different numerical schemes. Through these rigorous tests, we identify strengths and reveal important shortcomings of neural operators when handling complex nonlinear hyperbolic PDEs such as the Hughes model.

The numerical experiments demonstrate that although neural operators effectively approximate solutions under simpler conditions, their performance substantially deteriorates when confronted with realistic complexities inherent to pedestrian dynamics modeling. Thus, this study highlights the need for developing enhanced neural operator architectures or refined training strategies, supporting further advancement in machine learning for complex nonlinear hyperbolic systems.

The remainder of this paper is organized as follows. Section 2 presents an overview of the numerical methods used to solve the Hughes model and describes the procedures for generating training and testing datasets. In Section 3, we outline the experimental setup and report the results of applying state-of-the-art neural operators under a variety of challenging conditions. Finally, Section 4 concludes the paper and discusses potential directions for future research.

\section{Numerical schemes for Hughes model}
Due to the inherent complexity of the Hughes model, conventional numerical approaches often require fine discretization and substantial computational effort, for more details we refer to \cite{goatin2006awavefront, carrillo2016numerical, huang2009revisiting, liu2025multi} and references therein. To address these challenges, we employ both the wavefront tracking algorithm (WFT) \cite{goatin2006awavefront} and an adapted finite volume scheme (Godunov) \cite{godunov1959finite} to compute the solution. The wavefront tracking algorithm constructs piecewise-constant approximations and precisely tracks the evolution of discontinuities, making it particularly effective for capturing the hyperbolic structure of the governing equations with minimal numerical diffusion. Moreover, the Godunov scheme, situated within the finite-volume framework, solves localized Riemann problems at each cell interface, ensuring robust treatment of shocks and other nonlinear wave interactions. By coupling these two techniques, we gain a comprehensive perspective that not only facilitates cross-verification but also exploits the complementary strengths of each method—wavefront tracking’s refined handling of discontinuities and Godunov’s well-established stability in flux computations to produce more accurate and reliable solutions. In the following subsections, we explore the wavefront tracking algorithm in detail and subsequently discuss the Godunov scheme, providing further justification for employing both approaches in tandem.

\subsection[
  Wave‐front tracking scheme (Goatin et al.\ 2006)  
]{%
  Wave‐front tracking scheme \cite{goatin2006awavefront} 
}
\noindent
The wave--front tracking technique for classical scalar conservation laws constructs piecewise constant exact solutions of an approximated problem with piecewise constant initial data and a piecewise linear flux \(f_\epsilon\), where
$
f_\epsilon(\rho) = f(\rho) = \rho\,v(\rho),$
with \(v(\rho)\) the speed, on the mesh
$M_\epsilon = \{\rho_{i}^{\epsilon}\}_{i=0}^{2^\epsilon} \subset \Omega,$ with $\Omega = [0,1]$ as
\begin{equation}
M_\epsilon = 2^{-\epsilon}\mathbb{N} \cap [0,1].
\end{equation}
\textbf{Initialization.} Approximate the initial datum \(\rho_0\) by a piecewise constant function
\begin{equation}
\rho_0^{\epsilon}(x) = \sum_{j\in\mathbb{Z}} \rho_{0,j}^{\epsilon}\,\chi_{]x_{j-1},x_j]},
\end{equation}
with \(\rho_{0,j}^{\epsilon} \in M_\epsilon\), chosen so that
\begin{equation}
\lim_{\epsilon\to\infty}\|\rho_0^{\epsilon} - \rho_0\|_{L^1(\Omega)} = 0,\quad \text{and}\quad
\operatorname{TV}(\rho_0^{\epsilon}) \le \operatorname{TV}(\rho_0).
\end{equation}
Here, \(\operatorname{TV}(g)\) denotes the total variation of any bounded variation function \(g\) over the interval \(\Omega = [0,1]\), defined as
\[
\operatorname{TV}(g) = \sup_\mathcal{P} \sum_{j} \left| g(x_j) - g(x_{j-1}) \right|,
\]
where the supremum is taken over all finite partitions $\mathcal{P}$ of the interval. Intuitively, total variation measures the cumulative size of the jumps in g.
The points \(x_j's\) are not necessarily on a uniform grid; refinement is done on the density mesh \(M_\epsilon\). 

Set \(x_0 = \xi_0\), where \(\xi_0\) satisfies the cost balance equation at \(t=0\):
\begin{equation}
\int_{-1}^{\xi_0} c\bigl(\rho_0^{\epsilon}(y)\bigr)\,dy
=
\int_{\xi_0}^{1} c\bigl(\rho_0^{\epsilon}(y)\bigr)\,dy.
\end{equation}
In piecewise form:
\begin{equation}
\sum_{j\le 0} c(\rho_{0,j}^{\epsilon})(x_j - x_{j-1})
=
\sum_{j>0} c(\rho_{0,j}^{\epsilon})(x_j - x_{j-1}).
\end{equation}

For small \(t>0\), construct \((\rho^\epsilon,\xi_\epsilon)\) to approximate the coupled problem \eqref{eq:1.1} by solving local Riemann problems:

\noindent
\begin{minipage}[t]{0.48\textwidth}
\begin{equation}
\begin{cases}
\partial_t\rho + \partial_x \Bigl(\operatorname{sgn}(x-\xi_0)f^\epsilon(\rho)\Bigr)=0,\\[1mm]
\rho(0,x)=
\begin{cases}
\rho_{0,0}^{\epsilon} \quad \text{if } x<\xi_0,\\[1mm]
\rho_{0,1}^{\epsilon} \quad \text{if } x>\xi_0,
\end{cases}\\[1mm]
\dot{\xi}(\rho^+ - \rho^-) = \Psi[\rho],
\end{cases}
\end{equation}
\end{minipage}
\hfill
\begin{minipage}[t]{0.48\textwidth}
\begin{equation}
\begin{cases}
\partial_t\rho + \partial_x \Bigl(\operatorname{sgn}(x_j-\xi_0)f^\epsilon(\rho)\Bigr)=0,\\[1mm]
\rho(0,x)=
\begin{cases}
\rho_{0,j}^{\epsilon} \quad \text{if } x<x_j,\\[1mm]
\rho_{0,j+1}^{\epsilon} \quad \text{if } x>x_j,
\end{cases}\\[1mm]
j \neq 0.
\end{cases}
\end{equation}
\end{minipage}\\
where,
\begin{equation}
\left\{
\int_{\bar{\xi}+\delta}^{1} - \int_{-1}^{\bar{\xi}-\delta}
\right\}
\partial_t\bigl[c(\rho(t,y))\bigr]\,dy =: \Psi.
\end{equation}
Here, \(\bar{\xi}\) denotes the turning point at a fixed time, \(\delta > 0\) is a small constant that defines a buffer around \(\bar{\xi}\), and \(\Psi\) represents the net time variation of the cost across the buffer region.

Furthermore,
\begin{equation}
\rho^+(t) = \rho(t,\xi(t)^+) \quad \text{and} \quad \rho^-(t) = \rho(t,\xi(t)^-)
\end{equation}
are the right and left traces of \(\rho\) at \(x=\xi(t)\).

Solving the Riemann problems above replaces smooth rarefaction fronts by fans of constant values \(\rho_{1,j}^{\epsilon}\), \(j = l_1, \cdots, l_{N_{\epsilon}}\) such that
\begin{equation}
|\rho_{1,j}^{\epsilon}-\rho_{1,j-1}^{\epsilon}| = 2^{-\epsilon},
\end{equation}
separated by discontinuities with speeds
\begin{equation}
\lambda_{1,j} = -\,\operatorname{sgn}(\phi_x)\bigl(1-\rho_{1,j}^{\epsilon} - \rho_{1,j-1}^{\epsilon}\bigr).
\end{equation}

For the turning curve, use the Riemann solver given in Section 3 of \cite{goatin2006awavefront} (with \(\rho_M\) approximated by the closest point in \(M_\epsilon\)). The resulting new densities (if any) are inserted between the original states and relabeled as \(\rho_{1,j}^{\epsilon}\). Each pair of consecutive states is given a wave speed \(\lambda_{1,j}\) or \(\dot{\xi}_\epsilon\), with associated departure point \(x_j\). Hence, we get the wave trajectories as:
\begin{equation}
x_j(t) = x_j + \lambda_{1,j}\,t,\quad x_0(t)=\xi_\epsilon(t)=\xi_0+\dot{\xi}_\epsilon\,t,
\end{equation}
and to update the solution for small \(t>0\)
\begin{equation}
\rho^\epsilon(t,\cdot) = \sum_{j\in\mathbb{Z}} \rho_{1,j}^{\epsilon}\,\chi_{]x_{j-1}(t),\,x_j(t)]}.
\end{equation}

The piecewise constant approximate solution \(\rho^\epsilon\) constructed in the previous time step is extended until the first instance when either two waves interact or a wave encounters the turning curve \(\xi^\epsilon\). In both scenarios, the intermediate state is removed, and a new Riemann problem is formulated. Its solution is computed using the Reimann solver described in Section 3 of \cite{goatin2006awavefront} when the turning curve is reached, which permits further extension of \(\rho^\epsilon\) and \(\xi_\epsilon\) up to the subsequent interaction event. The absorbing boundary conditions are implemented by simply discarding any waves that strike the left or right boundaries. An illustration of an example solution can be seen in figure \ref{fig:wft_example_for_crowd}.
The MATLAB code used for the above scheme can be found at http://www-sop.inria.fr/members/Paola.Goatin/wft.html.

\subsection{Godunov scheme}
Godunov scheme is a numerical method used for solving hyperbolic conservation laws. This finite volume approach leverages approximate solutions of the Riemann problem at cell interfaces to capture discontinuities. 

\noindent
This section discusses the numerical solutions of~\eqref{eq:1.1} using the Godunov scheme \cite{godunov1959finite}. For a conservation law of type $u_t + F(u)_x = 0$, the Godunov flux is
\[
  h(u,v)
  =
  F\bigl(R(0;u,v)\bigr),
\]
where $R(0;u,v)$ is the solution of the Riemann problem with left and right states $u$ and $v$ evaluated at $x=0$. At each time step:
\begin{enumerate}
    \item Given $\rho(t=0,x)=\rho_0(x)$, determine the turning point $\xi(t)$ by using equation~\eqref{eq:cost_integral} using the method outlined in the Appendix (Section~\ref{sec:godunov algorithm}), rather than the Fast Sweeping algorithm discussed in \cite{zhao2005fast}. 

    \item Given $\xi(t)$, Solve the nonlinear conservation law~\eqref{eq:hughes_mod} using the Godunov scheme.
\end{enumerate}

Let $[-1,1]$ be divided into $N$ uniform cells $I_j=[x_{j-\tfrac{1}{2}},x_{j+\tfrac{1}{2}}]$ with centers $x_j=j\Delta x$, $\Delta x=2/N$ and $\Delta t^n =t^{n+1}-t^n $. 
The Godunov (numerical) fluxes are computed as:
\begin{equation}\label{eq:flux_right}
  q^{n-1}_{j+1/2} \;=\; F\bigl(\rho^{n-1}_j,\, \rho^{n-1}_{j+1})\bigr..
\end{equation}

The density is updated via
\begin{equation}\label{eq:density_update_right}
  \rho^n_j 
  = 
  \rho^{n-1}_j
  - 
  \frac{\Delta t^{n-1}}{\Delta x}\,\bigl(q^{n-1}_{j+1/2} - q^{n-1}_{j-1/2}\bigr).
\end{equation}

The flux function is given by:
\[
F(\rho^{n-1}_j, \rho^{n-1}_{j+1})=
\begin{cases}
\rho^{n-1}_{j}\,(1 - \rho^{n-1}_{j}), & \text{if } \rho^{n-1}_{j+1} \le \rho_{cr} \text{ and } \rho^{n-1}_{j} \le \rho_{cr}, \\[1mm]
q_{max},   & \text{if } \rho^{n-1}_{j+1} \le \rho_{cr} \text{ and } \rho^{n-1}_{j} > \rho_{cr}, \\[1mm]
\min\Bigl\{ \rho^{n-1}_{j+1}\,(1 - \rho^{n-1}_{j+1}),\, \rho^{n-1}_{j}\,(1 - \rho^{n-1}_{j}) \Bigr\}, & \text{if } \rho^{n-1}_{j+1} > \rho_{cr} \text{ and } k_{xup} \le \rho_{cr}, \\[1mm]
\rho^{n-1}_{j+1}\,(1 - \rho^{n-1}_{j+1}), & \text{if } \rho^{n-1}_{j+1} > \rho_{cr} \text{ and } \rho^{n-1}_{j} > \rho_{cr}.
\end{cases}
\]

\begin{figure}[htbp]
  \centering
  \begin{subfigure}[b]{0.31\textwidth}
    \centering
    \includegraphics[width=\textwidth,height=0.25\textheight,]{CROWD_2.pdf}
    \caption{Crowd Illustration}
    \label{fig:crowd_figure}
  \end{subfigure}
  \hspace{0.02\textwidth}
  \begin{subfigure}[b]{0.31\textwidth}
    \centering
    \includegraphics[width=\textwidth,height=0.31\textheight,keepaspectratio]{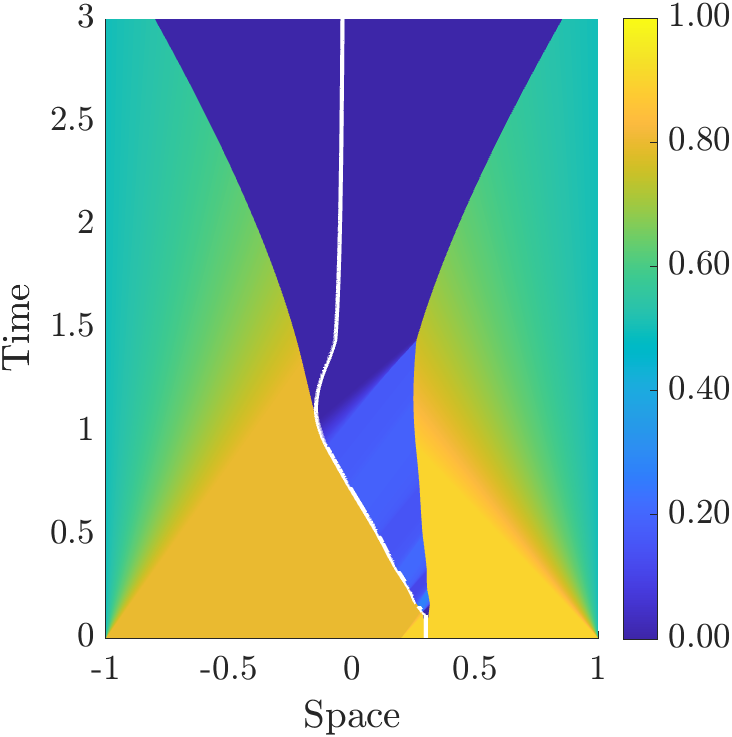}
    \caption{WFT algorithm simulation}
    \label{fig:wft_example_for_crowd}
  \end{subfigure}
  \hspace{0.02\textwidth}
  \begin{subfigure}[b]{0.31\textwidth}
    \centering
    \includegraphics[width=\textwidth,height=0.31\textheight,keepaspectratio]{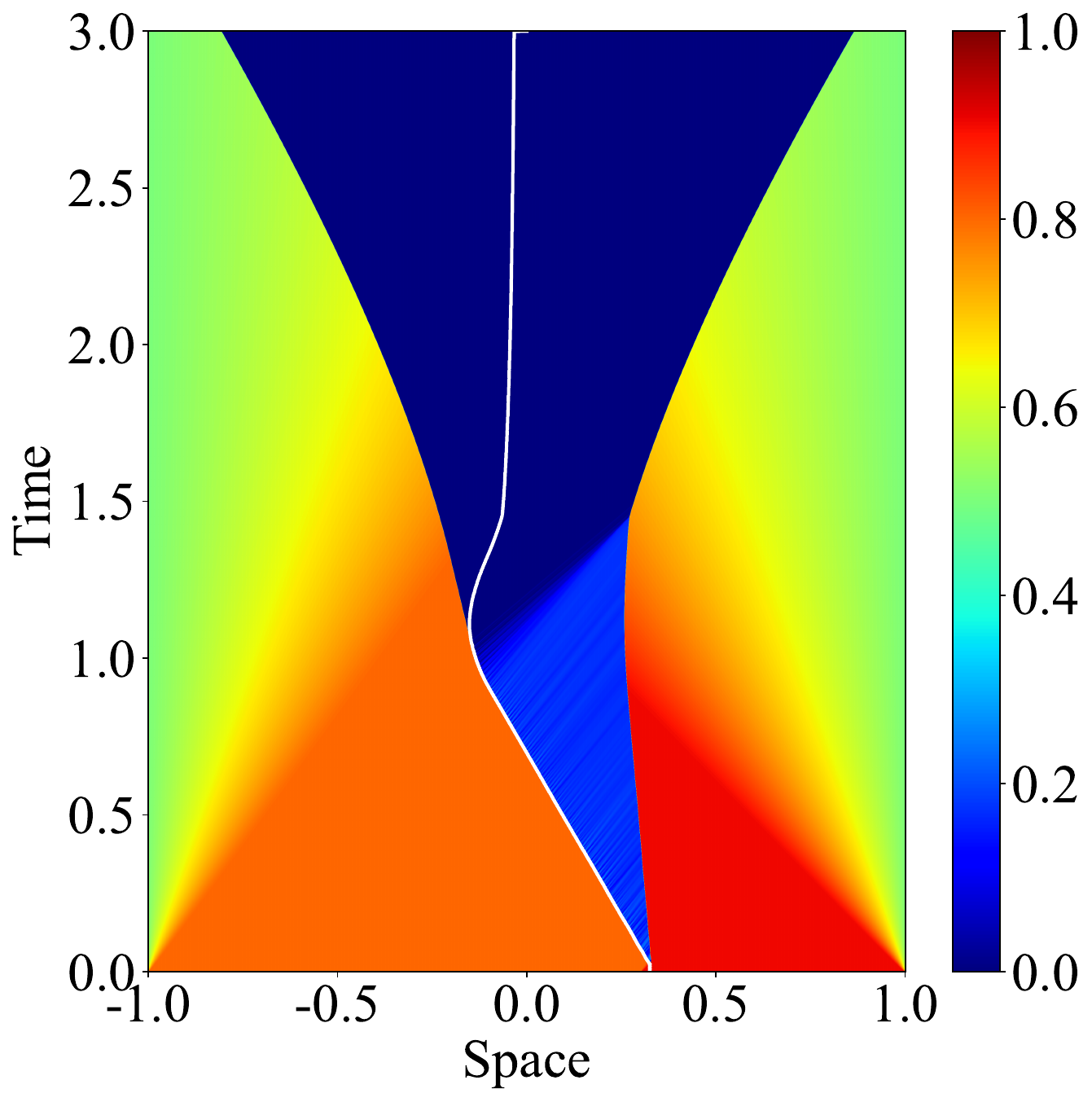}
    \caption{Godunov scheme simulation}
    \label{fig:godunovscheme_example}
  \end{subfigure}
  \caption{%
    \centering
    Illustration of Hughes pedestrian dynamics* \\[0.3ex]%
    {\footnotesize *High discretization used here for better visualization}%
  }%
\label{fig:crowd_dynamics_example}
\end{figure}

Here $q_{max} = 1$ and $\rho_{cr} = 1/2$. Now, update for each time step $t<T$. The resulting function \(\rho(x,t)\) is then the Godunov approximation of the Hughes model's solution. An illustration of an example solution can be seen in figure \ref{fig:godunovscheme_example}.

\subsection{Data generation}\label{data_generate}
The quality of the data is critical in the domain of operator learning. To rigorously assess the influence of data quality on the performance of neural operators, both training and testing datasets are generated using the numerical schemes detailed previously. In both approaches, the initial conditions (here, initial density profiles \(\rho(0,x)\)) and\slash or boundary conditions are provided as an input to compute the solution \(\rho(t,x)\) over the entire space-time grid. In this subsection, we present the procedure we follow to generate the different datasets that will be used for training and testing the neural operators.

\subsubsection{Generate initial conditions}\label{sec:algorithm-initial}
Two types of initial conditions are used to probe different dynamical regimes of the Hughes model. Firstly, the \emph{piecewise constant initial condition} which introduces discontinuities in the initial density profiles. These abrupt density jumps can create shocks and lead to complex interactions in pedestrian flow. Secondly, the \emph{Gaussian initial condition}, as they provide a smooth, continuous initial density profiles without discontinuities. It serves to isolate the effects of the boundary conditions on the evolution of the flow. The precise methodologies for constructing both the piecewise constant and Gaussian initial conditions are described below.\\

\noindent \textbf{Piecewise constant initial conditions:}\label{step_initial_condition}

The generation of the initial density profiles $\rho(t=0,x)$ is carried out in two main steps. First, $N+1$ density values are generated, where $N$ denotes the number of discontinuities in the initial condition. These values are then used to construct a piecewise constant (step) function over the spatial domain $[x_1, x_n]$. The first value, $k_1$, is sampled from a uniform distribution on the interval $[0.05, 0.95]$. For each subsequent index $j = 2, \dots, N+1$, the value $k_j$ is computed by adding a random increment $\delta_j \sim \mathcal{N}(0, 0.5)$ to $k_{j-1}$. Each $k_j$ is adjusted, if necessary, to ensure it satisfies the constraints $0 < k_j < 1$ and $k_j \neq k_{j-1}$.

\begin{figure}[htbp]
  \centering
  \begin{subfigure}[b]{0.31\textwidth}
    \centering
    \includegraphics[width=\textwidth,height=0.31\textheight,keepaspectratio]{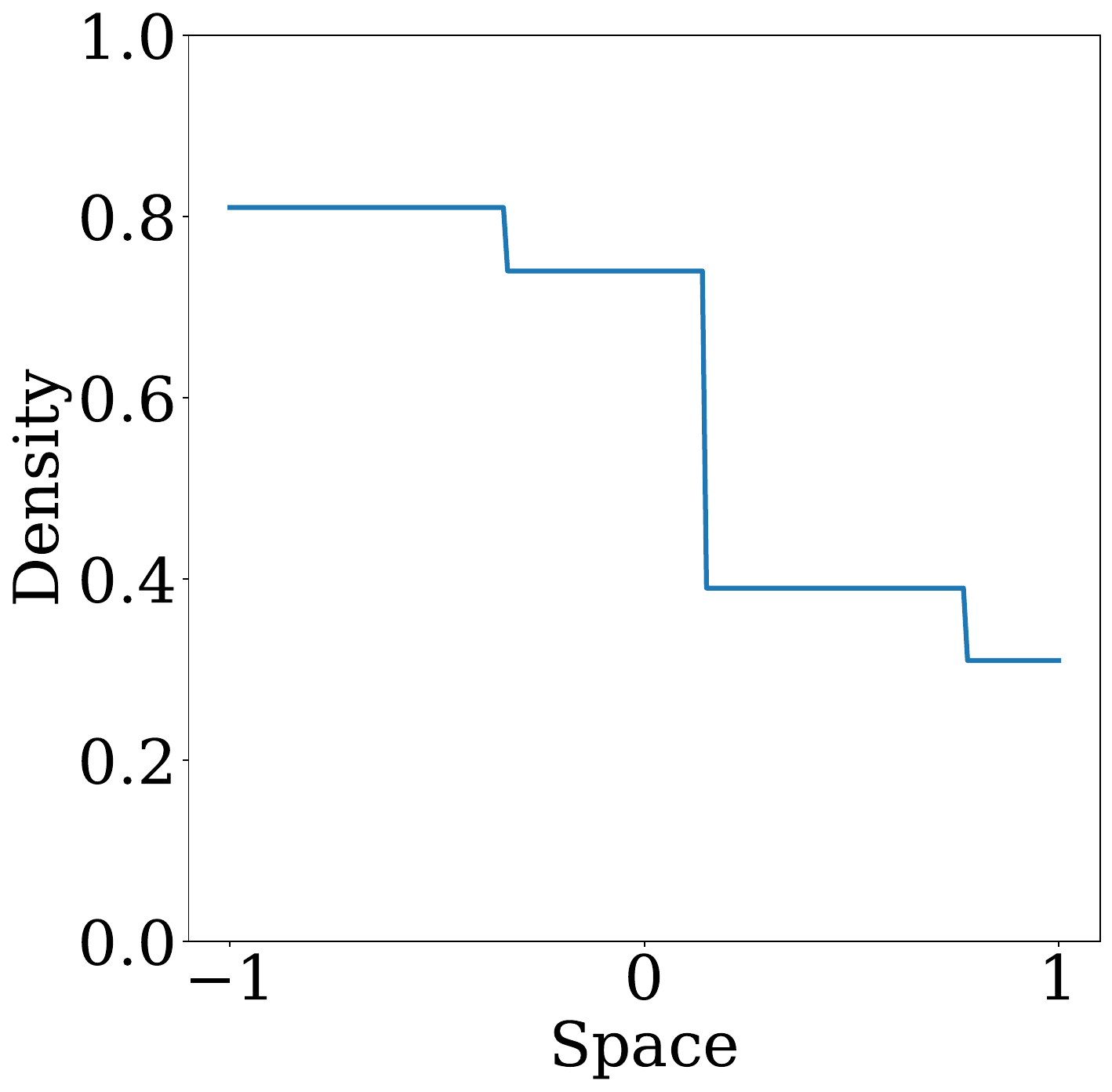}
    \caption{Three discontinuities}
    \label{fig:initial_cond_3step}
  \end{subfigure}
  \hspace{0.02\textwidth}
  \begin{subfigure}[b]{0.31\textwidth}
    \centering
    \includegraphics[width=\textwidth,height=0.31\textheight,keepaspectratio]{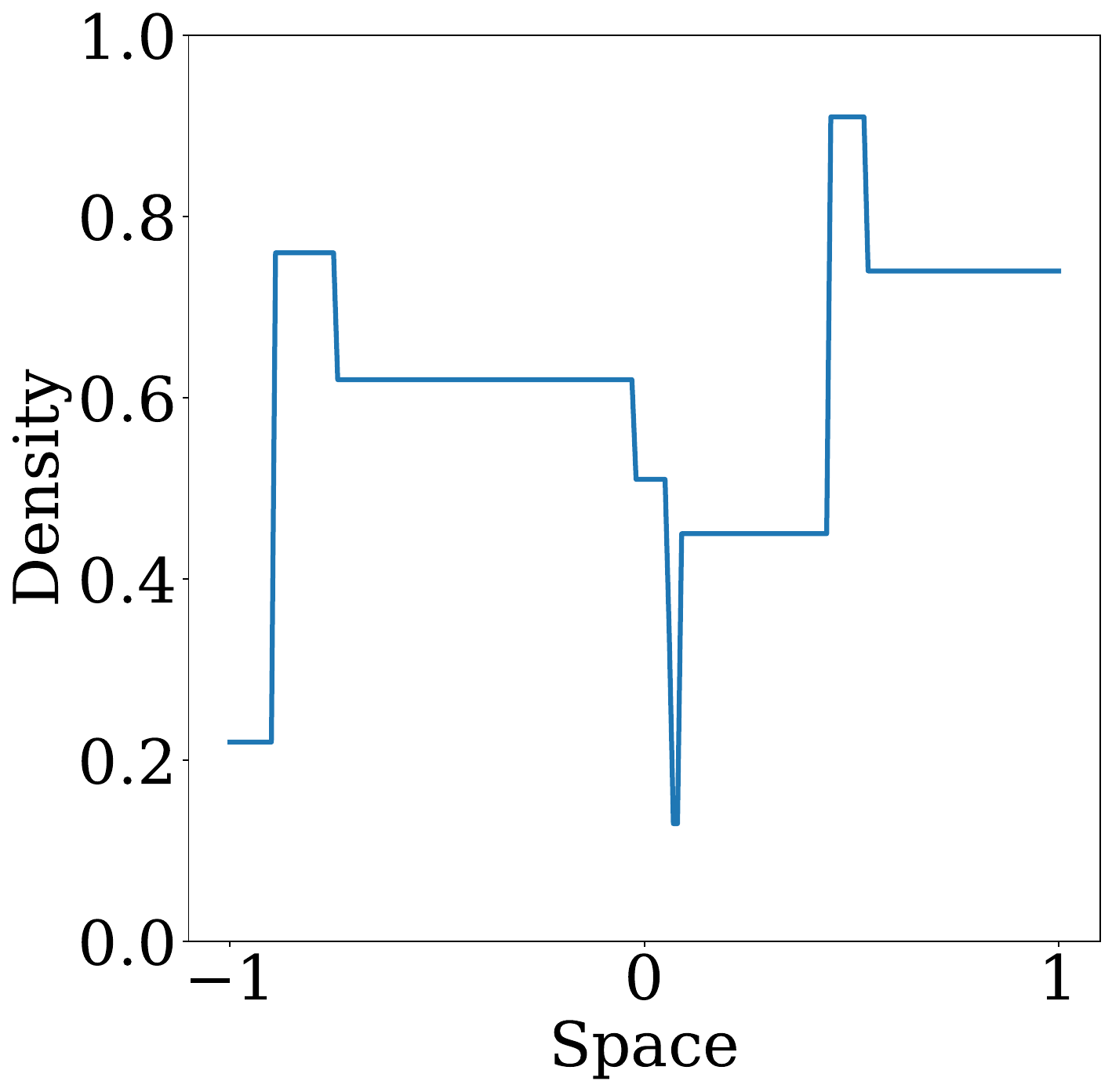}
    \caption{Seven discontinuities}
    \label{fig:initial_cond_7step}
  \end{subfigure}
  \hspace{0.02\textwidth}
  \begin{subfigure}[b]{0.31\textwidth}
    \centering
    \includegraphics[width=\textwidth,height=0.31\textheight,keepaspectratio]{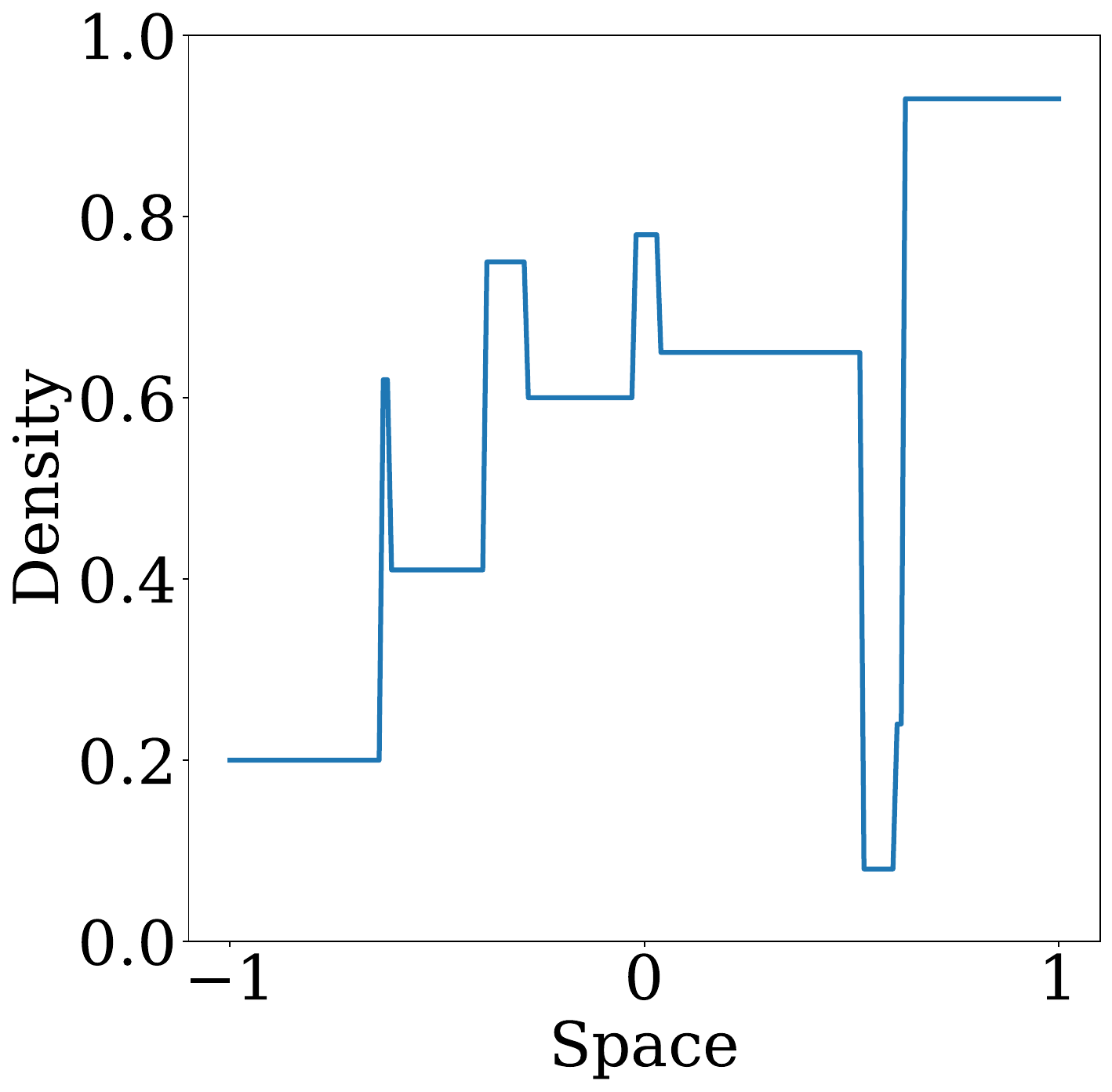}
    \caption{Nine discontinuities}
    \label{fig:initial_cond_9step}
  \end{subfigure}
  \caption{Example data for initial conditions with multiple discontinuities}
  \label{fig:initial_cond_ex}
\end{figure}

In the second step, $N$ unique positions $s_i$ are sampled uniformly from $[x_1, x_n]$. These positions are then sorted in ascending order, partitioning the spatial domain into $N+1$ contiguous regions:
\[
r_1 = [x_1, s_1), \quad 
r_2 = [s_1, s_2), \quad
\dots, \quad
r_{N} = [s_{N-1}, s_N), \quad
r_{N+1} = [s_N, x_n].
\]
Each density value $k_j$ is assigned to its corresponding sub-interval, producing constant-density regions: $\rho_{r_1} = k_1, \ \rho_{r_2} = k_2, \ \dots, \ \rho_{r_{N+1}} = k_{N+1}$. This procedure is applied to generate the datasets used in both Problem I  and Problem II in Subsection ~\ref{problems123}. Representative examples of these initial conditions are provided in Figure~\ref{fig:initial_cond_ex}. The hyperparameters used in the Godunov scheme and wavefront-tracking algorithm are listed in Table~\ref{tab:simulation_parameters}. Meanwhile, Figure~\ref{fig:combined_examples} illustrates how the same initial condition evolves under both solvers, highlighting the resulting density profile $\rho(t,x)$.\\

\noindent \textbf{Gaussian initial condition: }\label{Gaussian_initial_condition}

The spatial domain \( [x_1, x_n] \) is initialized with a continuous Gaussian pulse to ensure a smooth initial density profile. The profile is defined by a Gaussian function whose parameters are randomly sampled as follows:
\[
\mu \sim \mathcal{U}(0, x_n), \quad \sigma \sim \mathcal{U}\left(0, \frac{x_n}{2}\right),
\]
where \(\mu\) denotes the mean and \(\sigma\) the standard deviation of the Gaussian. The initial density is then given by
\[
\rho(t=0,x) = \exp\left(-\frac{(x - \mu)^2}{2\sigma^2}\right).
\]
This procedure is used to generate the initial density profiles for the data used in Problem III in Subsection~\ref{problems123}.

\begin{table}[htb]
\centering
\caption{Simulation parameters for Godunov scheme and WFT algorithm}
\label{tab:simulation_parameters}
\begin{tabular}{lcc}
\toprule
\textbf{Parameter} & \textbf{Godunov Scheme} & \textbf{WFT Algorithm} \\
\midrule
Domain \(x\) & \([-1, 1]\) & \([-1, 1]\) \\
Time interval \(t\) & \([0, 3]\) & \([0, 3]\) \\
Spatial resolution \(\Delta x\) & \(1/500\) & \(2/201\) \\
Temporal resolution \(\Delta t\) & \(1/600\) & \ \(1/2000\) \\
Maximum density \(\rho_{\max}\) & \(1\) & 1 \\
Maximum velocity \(v_{\max}\) & \(1\) & 1 \\
Critical density \(\rho_{\text{cr}}\) & \(1/2\) & -- \\
Maximum flow \(q_{\max}\) & \(1/4\) & -- \\
Mesh size \(\Delta\rho\) & -- & \(1/250\) \\
\bottomrule
\end{tabular}
\end{table}

\subsubsection{Generate boundary conditions}\label{sec:algorithm-boundary}
The flow at the boundaries is regulated via the potential function \(\phi(t,x)\) (see equation~\eqref{eq:1.3b}), with pedestrians naturally moving from regions of high to low potential. In this study, we consider three distinct scenarios for the boundary conditions. In Problem I in Subsection~\ref{problems123}, pedestrians are allowed to exit the domain freely by setting:
\begin{subequations}\label{eq:bc_1}
\begin{align}
& \phi(t,-1) =\phi_{bc1}(t) =0, \qquad    \phi(t,1) = \phi_{bc2}(t) =0.
\end{align}
\end{subequations}
In the second scenario in Problem II in Subsection~\ref{problems123}, high potential values are imposed near \(x=\pm 1\) to create a repulsive effect that directs pedestrians away from the exits and into the interior of the domain, modeling their behavior at the closed exits.
\begin{subequations}\label{eq:bc_2}
\begin{align}
& \phi(t,-1) =\phi_{bc1}(t) = 10000, \qquad    \phi(t,1) = \phi_{bc2}(t) =10000. 
\end{align}
\end{subequations}
Finally, a third scenario in Problem III in Subsection~\ref{problems123}, features time-varying boundary conditions, where the value of \(\phi(t,x)\) at \(x= -1\) is varied with time as follows. 
\\

\begin{subequations}\label{eq:bc_3}
\begin{align}
\phi(t,-1) = \phi_{bc1}(t) &= 
\begin{cases}
10000, & 0 \le t < t_1,\quad t_1 \sim \mathcal{U}(0.08, 0.5),\\[1mm]
0, & t_1 \le t < t_2,\quad t_2 \sim \mathcal{U}(0.8, 1),\\[1mm]
10000, & t_2 \le t < t_3,\quad t_3 \sim \mathcal{U}(1.4, 2.2),\\[1mm]
0, & t_3 \le t < t_4,\quad t_4 \sim \mathcal{U}(2.5, 2.75),\\[1mm]
10000, & t_4 \le t < t_5,\quad t_5 \sim \mathcal{U}(2.8, 3.33),\\[1mm]
0, & t_5 \le t < 5,
\end{cases} \label{eq:bc_potentials_a} \\[2mm]
\phi(t,1) = \phi_{bc2}(t) &= 0,\quad \forall\, t \in [0,5]. \notag
\end{align}
\end{subequations}

\subsubsection{Generate dataset for various test scenarios}\label{data_gen}

\begin{figure}[htbp]
  \centering
  \begin{subfigure}[b]{0.48\textwidth}
    \centering
    \includegraphics[width=\textwidth, height=0.3\textheight, keepaspectratio]{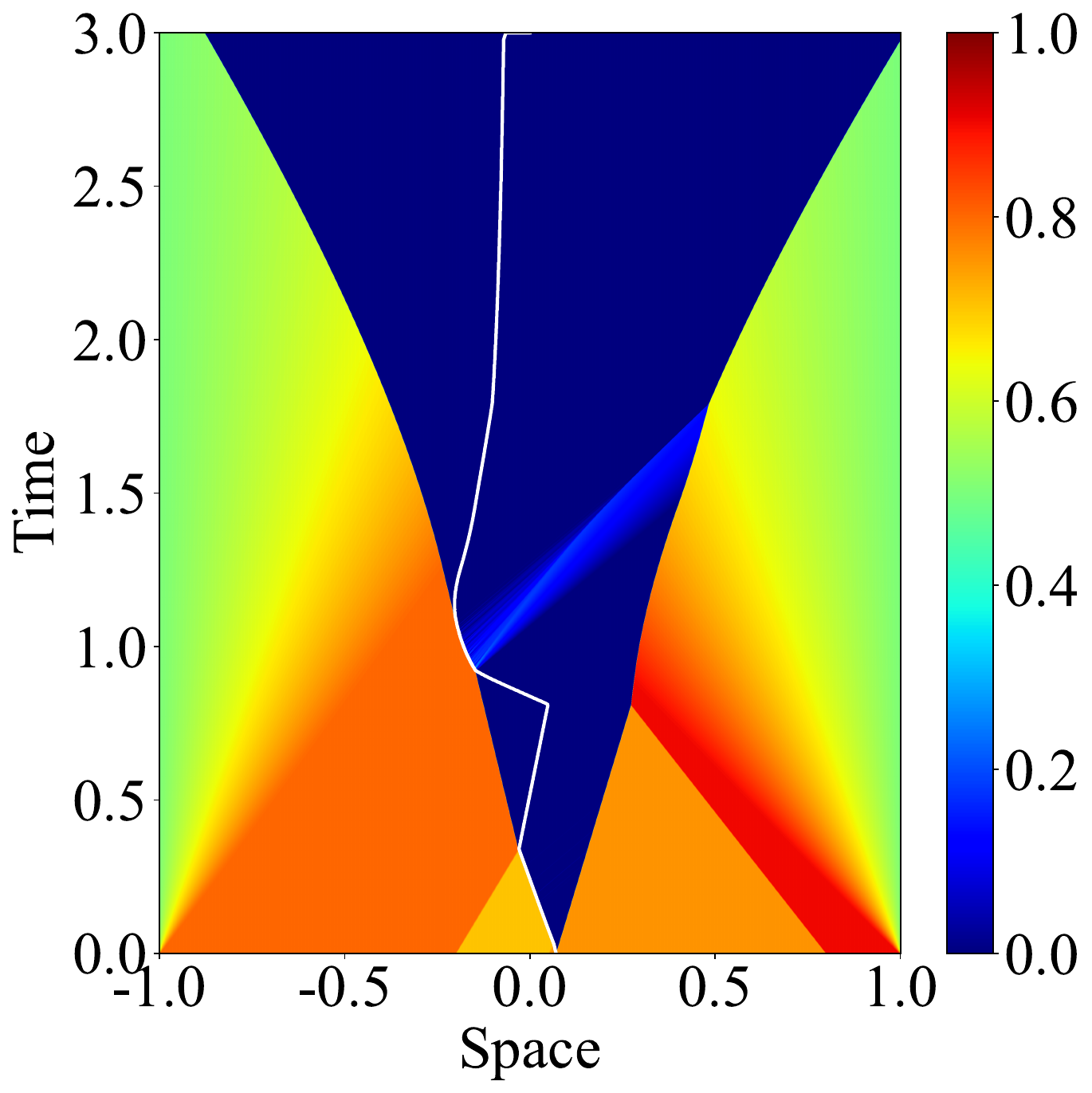}
    \caption{Data Generated by Godunov Scheme}
    \label{fig:godunov_example}
  \end{subfigure}
  \hspace{0.02\textwidth}
  \begin{subfigure}[b]{0.48\textwidth}
    \centering
    \includegraphics[width=\textwidth, height=0.3\textheight, keepaspectratio]{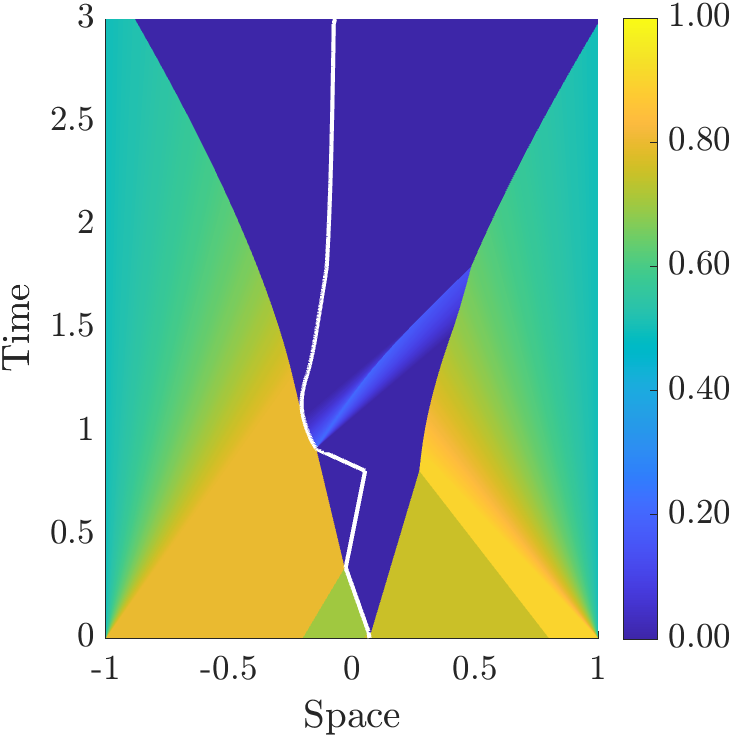}
    \caption{Data Generated by WFT Algorithm}
    \label{fig:wft_example}
  \end{subfigure}  
    \caption{%
    \centering
    Example data for Godunov and WFT scheme, with same initial condition* \\[0.3ex]%
    {\footnotesize *High discretization used here for better visualization}%
  }%
  \label{fig:combined_examples}
\end{figure}

The methodology used to generate the dataset for neural operators for different test scenarios is as follows: For Problem I given in Subsection \ref{problems123}, for both the numerical methods, two datasets are produced, classified as \textit{easy} and \textit{complex}. \textit{Easy} samples are defined as those for which the difference between the maximum and minimum turning point ($ \Delta \xi = \xi_{\text{max}} - \xi_{\text{min}}$) is less than 0.3, having up to three discontinuities in their initial density profile ($N\leq 3$, where, $N$ = number of discontinuities in the initial density profile). In contrast, \textit{complex} samples are characterized by a turning point difference greater than 0.3 ($ \Delta \xi > 0.3 $), having one to ten discontinuities in their initial density profile ($N \leq 10$). This choice of \( \Delta \xi \) is supported by our observation that as \( \Delta \xi \) increases, the system becomes more complex. Consequently, \textit{complex} samples display a higher degree of variability in turning points and more pronounced discontinuities in their initial density profile, leading to more intricate patterns in the solution $\rho(t,x)$. 
 For the \textit{easy} samples ($ \Delta \xi < 0.3 \text{ and } N\leq3$), $\rho(t=0,x)$ follows the same distribution with parameters as shown in section \ref{sec:algorithm-initial} for both the methods, which means, the data generated from the two numerical methods may not be exactly similar but share a common distribution. For the \textit{complex} samples ($ \Delta \xi > 0.3 \text{ and } N \leq 10$), $\rho(t=0,x)$ are generated once. Then, these same initial conditions are used to generate solution $\rho(t,x)$, using both the numerical methods to get a fair comparison. The boundary conditions used for both \textit{easy} and \textit{complex} samples are same and are given by equation (\ref{eq:bc_1}). Figure \ref{fig:combined_examples} shows an example of a sample generated by the Godunov scheme and the WFT algorithm for the same 
 $\rho(t=0,x)$. 
 
For Problem II described in Subsection~\ref{problems123}, samples feature initial conditions with up to three discontinuities only. The corresponding boundary conditions for this case are different from Problem I and are provided in equation~(\ref{eq:bc_2}). Moreover, the data for this scenario is generated exclusively using the Godunov scheme.

For Problem III described in Subsection~\ref{problems123}, Gaussian initial conditions are employed. Specifically, the initial density profile \(\rho(0,x)\) follows a Gaussian distribution (see Subsection~\ref{Gaussian_initial_condition}). This problem is further subdivided into two cases based on the boundary conditions: (i) boundary conditions as given in equation~(\ref{eq:bc_1}), and (ii) boundary conditions as given in equation~(\ref{eq:bc_3}). The data for both cases are generated exclusively using the Godunov scheme.

\subsubsection{Data preprocessing} \label{dataset}

The dataset for the Godunov scheme and the WFT algorithm is generated using the hyperparameters shown in Table \ref{tab:simulation_parameters}. Each dataset is down sampled to reduce computational cost while preserving essential solution features. For the Godunov scheme, each raw sample is initially generated with a spatiotemporal resolution of \((x=1000,t=1800)\), down sampled to a shape of \((x=200,t=50)\), which is used for training the FNO and WNO. 

In contrast, the sample generated via the WFT algorithm is of shape $(x=201,t=6000)$, which is then down sampled to $(x=201,t=50)$. This choice of generating data at $x=201$ is deliberate, as WFT inherently produces stair-step data due to its discontinuous structure, and retaining full spatial resolution is necessary to accurately capture the locations of discontinuities at each time step. Since using $201$ points between $[-1,1]$ gives a discretization of $0.01$, it becomes easier to round the values to 2 decimal places to preserve the location of discontinuities.

For the multi-wavelet neural operator, which requires square input domains due to the structure of multi-wavelet transforms, the samples from both numerical methods are uniformly down sampled to a resolution of \((128, 128)\).

\begin{figure}[htbp]
  \centering
  \begin{subfigure}[b]{0.31\textwidth}
    \centering
    \includegraphics[width=\textwidth,height=0.31\textheight,keepaspectratio]{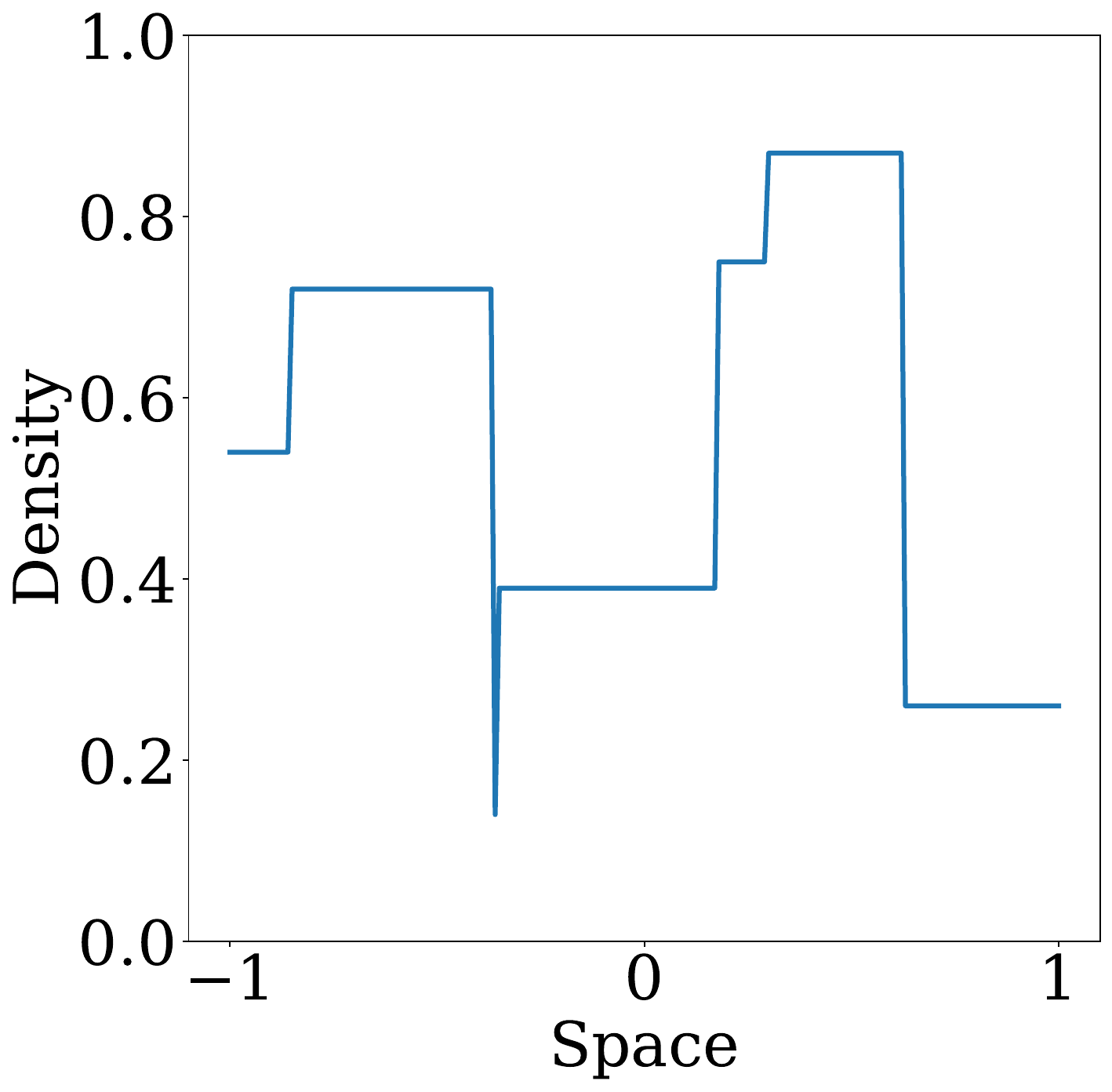}
    \caption{Initial Condition}
    \label{fig:initialcond}
  \end{subfigure}
  \hspace{0.02\textwidth}
  \begin{subfigure}[b]{0.31\textwidth}
    \centering
    \includegraphics[width=\textwidth,height=0.31\textheight,keepaspectratio]{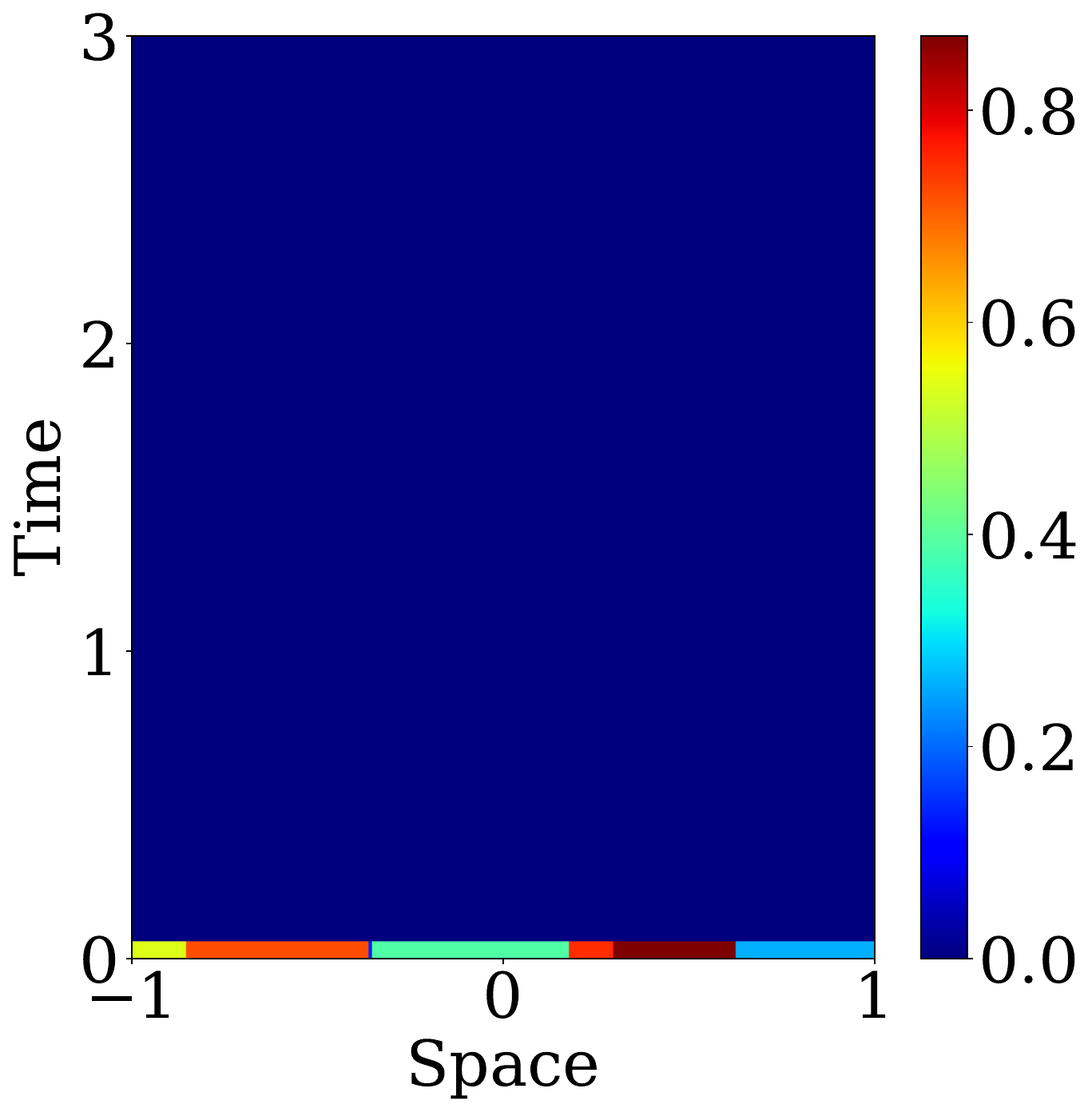}
    \caption{Training Input (X)}
    \label{fig:Training_input}
  \end{subfigure}
  \hspace{0.02\textwidth}
  \begin{subfigure}[b]{0.31\textwidth}
    \centering
    \includegraphics[width=\textwidth,height=0.31\textheight,keepaspectratio]{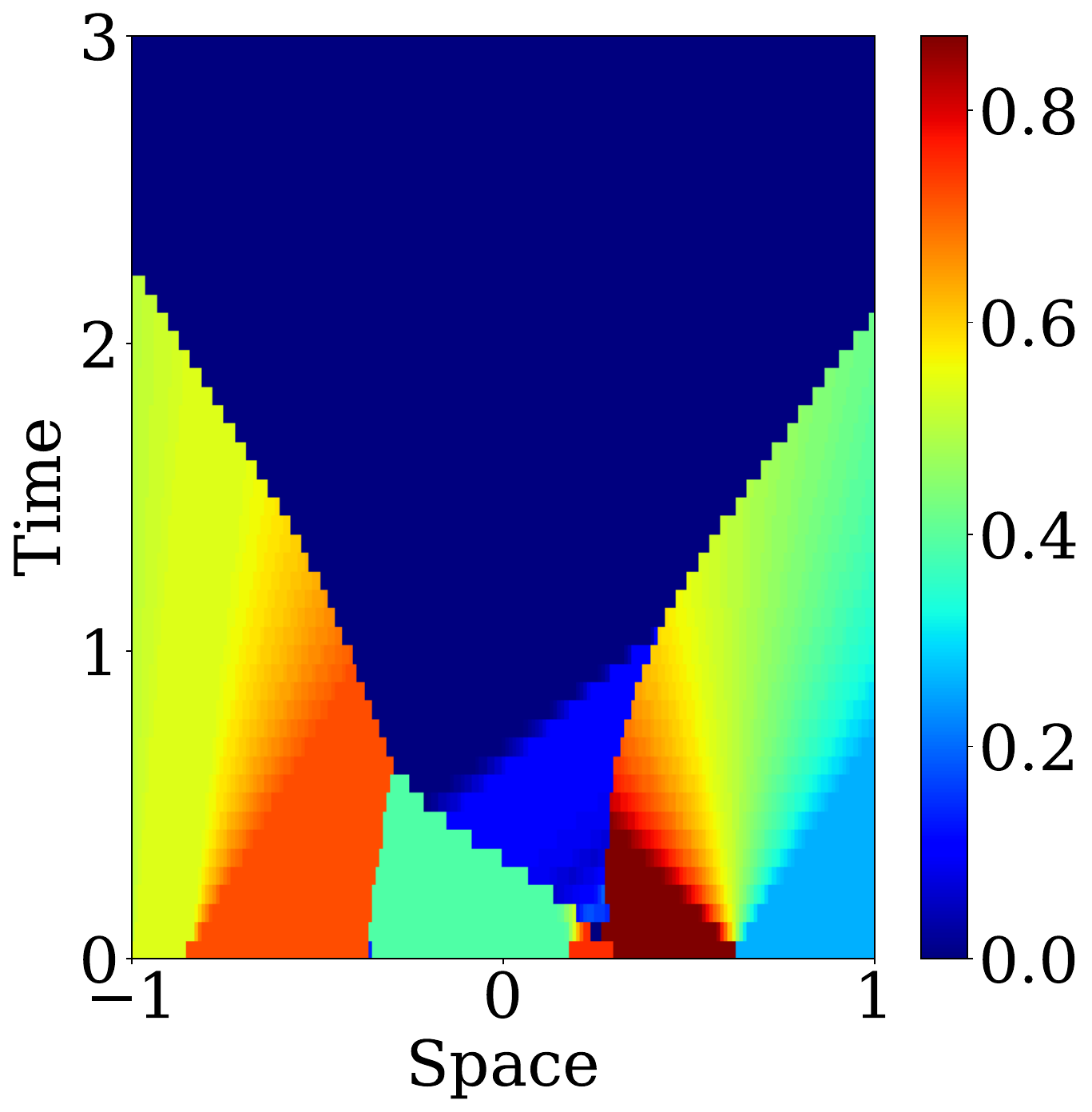}
    \caption{Target Output (Y)}
    \label{fig:train_out}
  \end{subfigure}
  \caption{Training data for initial value problem}
  \label{fig:training_data}
\end{figure}

\begin{figure}[htbp]
  \centering
  \begin{subfigure}[b]{0.48\textwidth}
    \centering
    \includegraphics[width=\textwidth, height=0.3\textheight, keepaspectratio]{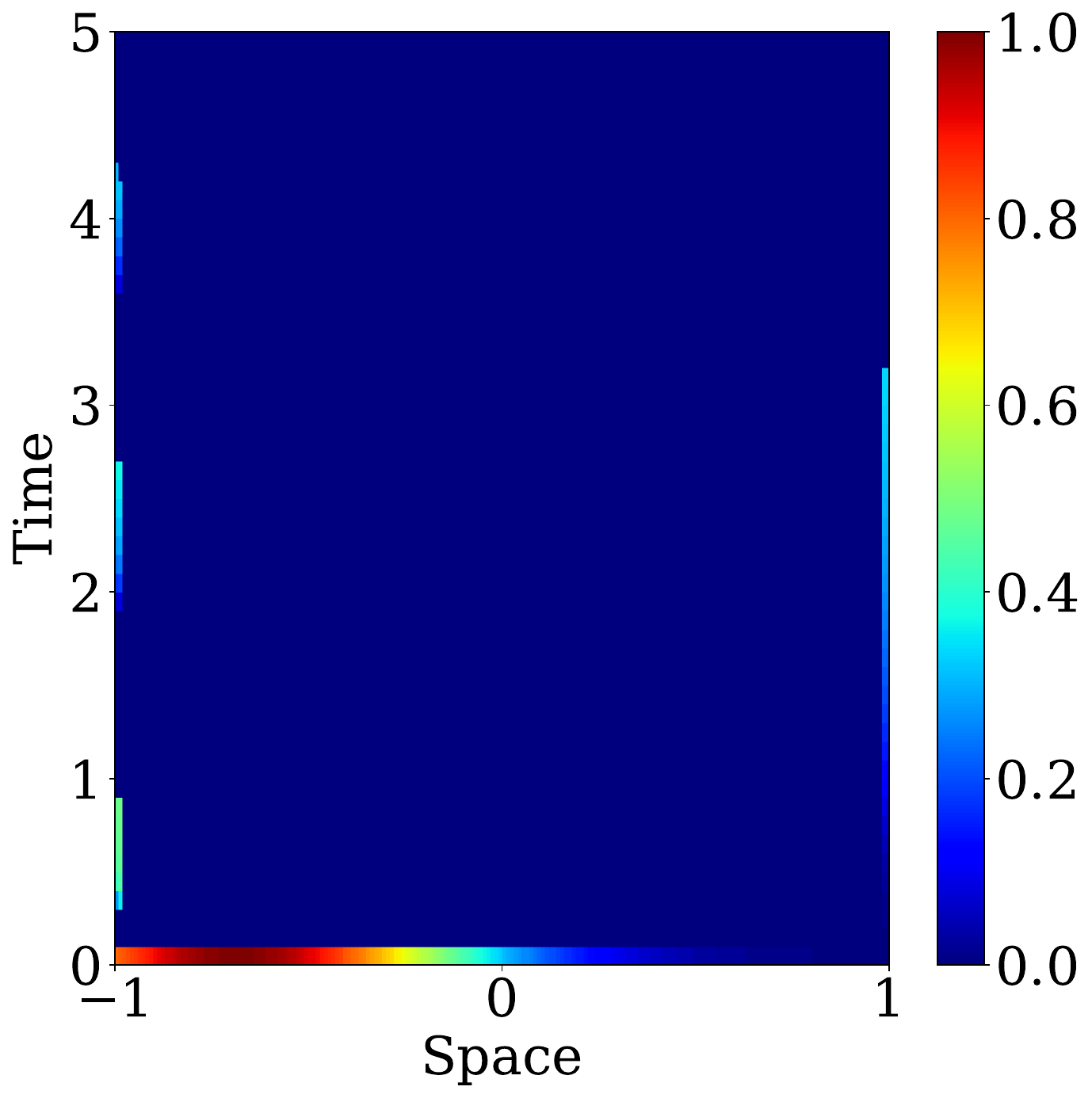}
    \caption{Training Input (X)}
    \label{fig:Training Input BC II}
  \end{subfigure}
  \hspace{0.02\textwidth}
  \begin{subfigure}[b]{0.48\textwidth}
    \centering
    \includegraphics[width=\textwidth, height=0.3\textheight, keepaspectratio]{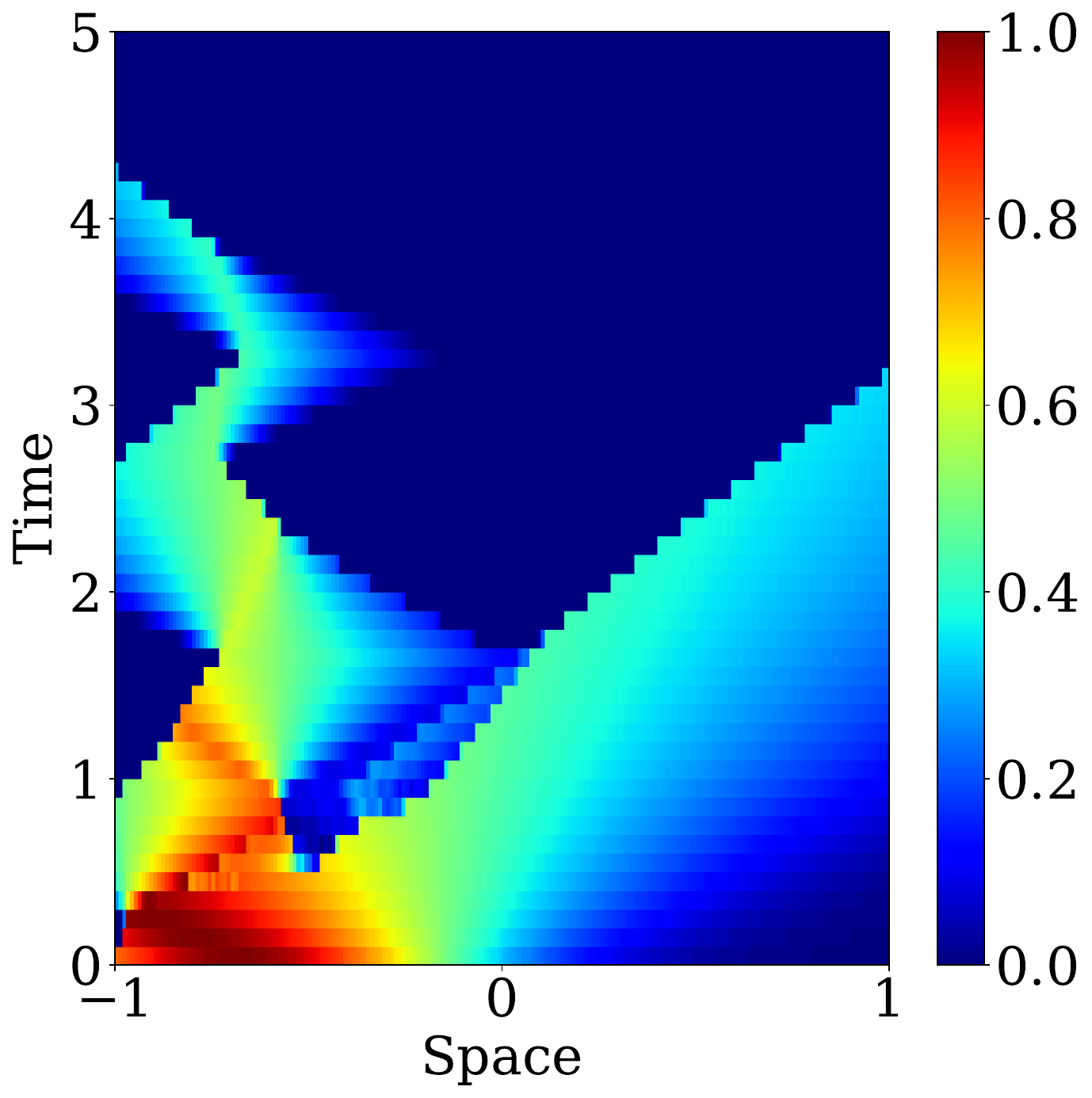}
    \caption{Target Output (Y)}
    \label{fig:Training Output_BC II}
  \end{subfigure}
  \caption{Training data for mixed initial-boundary value problem}
  \label{fig:BC II dataset}
\end{figure}

Each sample is divided into two components: the training input \(X\) and the target output \(Y\). For problems I and II in Subsection \ref{problems123}, the input \(X\) consists of the initial condition \(\rho(0,x)\) embedded into a tensor with the same dimensions as the full spatiotemporal grid, with all values for \(t > 0\) set to zero (see Figure~\ref{fig:Training_input}). Problem III, described in Subsection~\ref{problems123}, is divided into two subproblems. In the first sub-problem, the initial value problem, the input \(X\) is constructed in the same manner as in the other cases (see Figure~\ref{fig:Training_input}). In the second sub-problem, the mixed initial boundary value problem, the input \(X\) includes both the initial condition and the boundary conditions, while all other values of \(\rho(t,x)\) for \(t>0\) and \(-1<x<1\) are set to zero (see Figure~\ref{fig:Training Input BC II}). In every case, the target output \(Y\) comprises the complete solution trajectory \(\rho(t,x)\) (see Figures~\ref{fig:train_out} and~\ref{fig:Training Output_BC II}).

\section{Experiments and results} 

This section presents the experimental setup and evaluation of the neural operators across the different test scenarios. The first subsection defines the metric used for evaluation of the neural operators. Subsequent subsections detail the quantitative and qualitative results obtained from the experiments.

\subsection{Model evaluation}
For both training and evaluation, the relative \(L_2\) loss is used as the primary error metric. This loss function measures the deviation between the predicted solution and the ground truth, normalized by the magnitude of the ground truth itself. It is defined as:

\begin{equation}
\text{Data Loss} = \min_{\Theta} \frac{1}{|\mathcal{D}|} \sum_{i \in \mathcal{I}} \left\| \mathcal{L}(G_{\Theta}(\mathbf{X}(i)), \mathbf{Y}(i)) \right\|_2^2,
\end{equation}

where \(\Theta\) denotes the trainable parameters of the neural operator, \(\mathcal{D}\) is the dataset, \(\mathcal{I}\) is the index set of training samples, and \(G_{\Theta}(\mathbf{X}(i))\) is the model's prediction for the input \(\mathbf{X}(i)\). The function \(\mathcal{L}(\cdot, \cdot)\) computes the relative error between the predicted and true solution fields:

\begin{equation}
\mathcal{L}(\hat{\mathbf{Y}}, \mathbf{Y}) = \frac{\left\| \hat{\mathbf{Y}} - \mathbf{Y} \right\|_2}{\left\| \mathbf{Y} \right\|_2},
\end{equation}

where \(\hat{\mathbf{Y}}\) is the predicted solution and \(\mathbf{Y}\) is the ground truth.

By normalizing the error, this metric ensures that all samples contribute equally to the overall loss, regardless of their absolute scale. It also allows for meaningful comparison of model performance across datasets with different dynamics or initial conditions. This relative error metric is used consistently across all neural operators (Fourier neural operator (FNO), wavelet neural operator (WNO), and multiwavelet neural operator (MWNO)) and all data variants (Godunov/WFT, easy/complex), allowing for fair and robust comparison of performance across architectures and scenarios.

\subsection{Total variation-based diffusion metrics}

While the relative \(L^2\) error captures global approximation accuracy, it does not reflect how well the model preserves physical features such as discontinuities. To address this, we introduce secondary metrics based on the total variation (TV) of the solution, which are sensitive particularly to diffusion. Total variation provides a global measure of how much the solution varies across space, making it informative for transport-dominated problems.

Given a fixed sample and a solution profile \( \rho_j^n \) at time step \( n \in \{0, \dots, T \}  \) and spatial grid index \( j \), the discrete total variation is defined as:
\[
\operatorname{TV}(u^n) = \sum_{j=1}^{N-1} \left| \rho_{j+1}^n - \rho_j^n \right|.
\]
For any sample \( m \in \{1, \dots, M \} \) and time step \( n \), we denote the TV difference by
\[
\delta_{\mathrm{TV}}^{n,m} = \operatorname{TV}(\rho^{n,m}) - \operatorname{TV}(\hat{\rho}^{n,m})
,
\]
where \( \rho^{n,m} \) and \( \hat{\rho}^{n,m} \) denote the ground truth and predicted solutions, respectively. Positive values of \(\delta_{\mathrm{TV}}^{n,m}\) indicate that the predicted solution has less total variation than the reference, which may be due to smoothing of discontinuities, damping of valid oscillations, or flattening of variable regions. Negative values, on the other hand, may reflect excess variation introduced by oscillations or artificial steepness (over-sharpening).

While the signed metric captures both under- and over-variation, our focus is on identifying smoothing behavior. Therefore, we compute the "clipped TV" loss as:
\[
\overline{\delta}_{\mathrm{TV}}^{+} = \frac{1}{MT} \sum_{m=1}^{M} \sum_{n=1}^{T} \max\left(0, \delta_{\mathrm{TV}}^{n,m} \right),
\]
which isolates cases where the predicted solution underestimates total variation. This provides a more targeted view of potential diffusive effects in the model.

\subsection{Numerical experiments and results}\label{problems123}
To systematically evaluate the performance and generalization capabilities of neural operators, a series of experiments are designed. These experiments involve various initial and boundary conditions. The experimental setup consists of the following three problems.
\subsubsection{Problem  I} \label{problem1}
We begin the evaluation by testing FNO, WNO and MWNO on the datasets described in Subsection \ref{dataset} (see also Figure \ref{fig:training_data}). Each of these three neural operators are trained and tested on both \textit{easy} and \textit{complex} datasets generated independently by the Godunov scheme and the Wavefront Tracking (WFT) algorithm. This setup enables a comprehensive comparison of operator performance across varying levels of difficulty and over different numerical solvers. 

The primary goal is to assess not only how each operator handles increasing complexity in the initial conditions, but also whether there is any marked difference in learning patterns between data produced by the Godunov scheme and the WFT algorithm. The hyperparameter for all three operators are summarized in Table~\ref{tab:NOs hyperparams}.

\begin{figure}[htbp]
  \centering
  \begin{subfigure}[b]{0.485\textwidth}
    \centering
    \includegraphics[width=\textwidth,height=0.31\textheight,keepaspectratio]{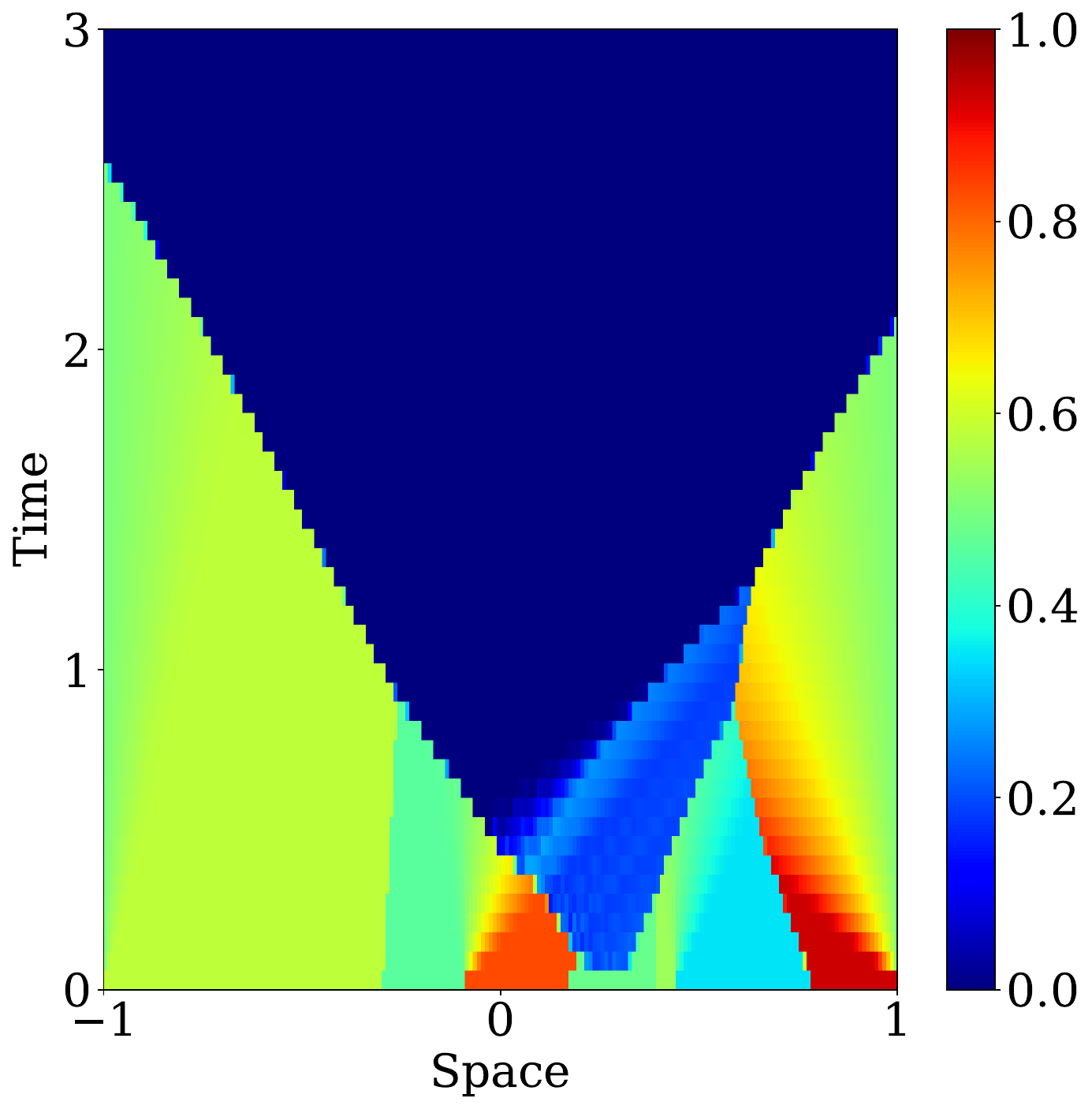}
    \caption{Real Data}
    \label{fig:real-data-gd-fno-sec7.1}
  \end{subfigure}
  \hspace{0.01\textwidth}
  \begin{subfigure}[b]{0.485\textwidth}
    \centering
    \includegraphics[width=\textwidth,height=0.31\textheight,keepaspectratio]{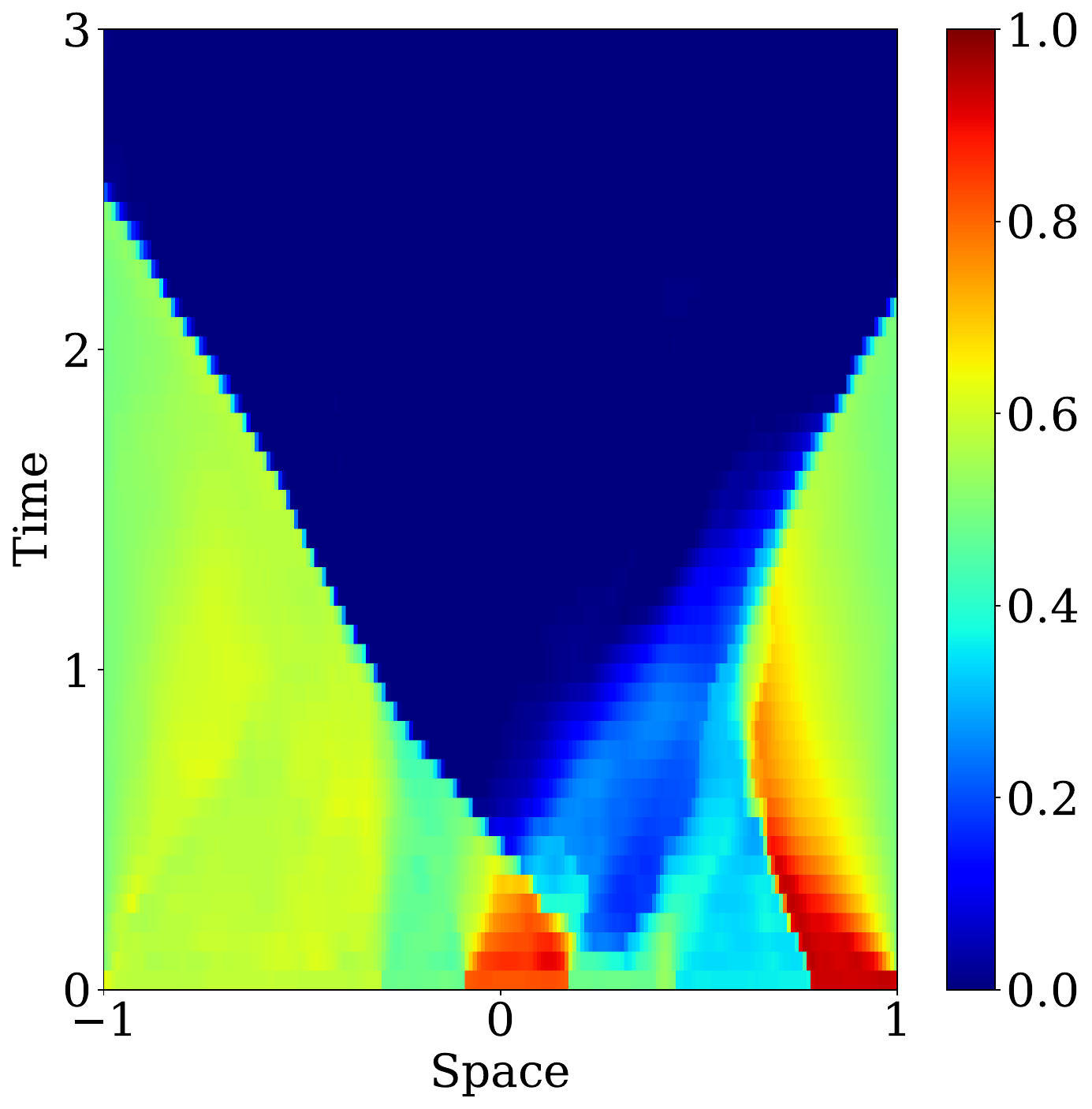}
    \caption{Predicted Data}
    \label{fig:predicted-data-gd-fno-sec7.1}
  \end{subfigure}
  \hspace{0.02\textwidth}
  \caption{Prediction results for data with six discontinuities in initial condition, generated using Godunov Scheme and trained on FNO }
  \label{fig:gd-fno-sect7.1}
\end{figure}

\begin{figure}[htbp]
  \centering
  \begin{subfigure}[b]{0.485\textwidth}
    \centering
    \includegraphics[width=\textwidth,height=0.31\textheight,keepaspectratio]{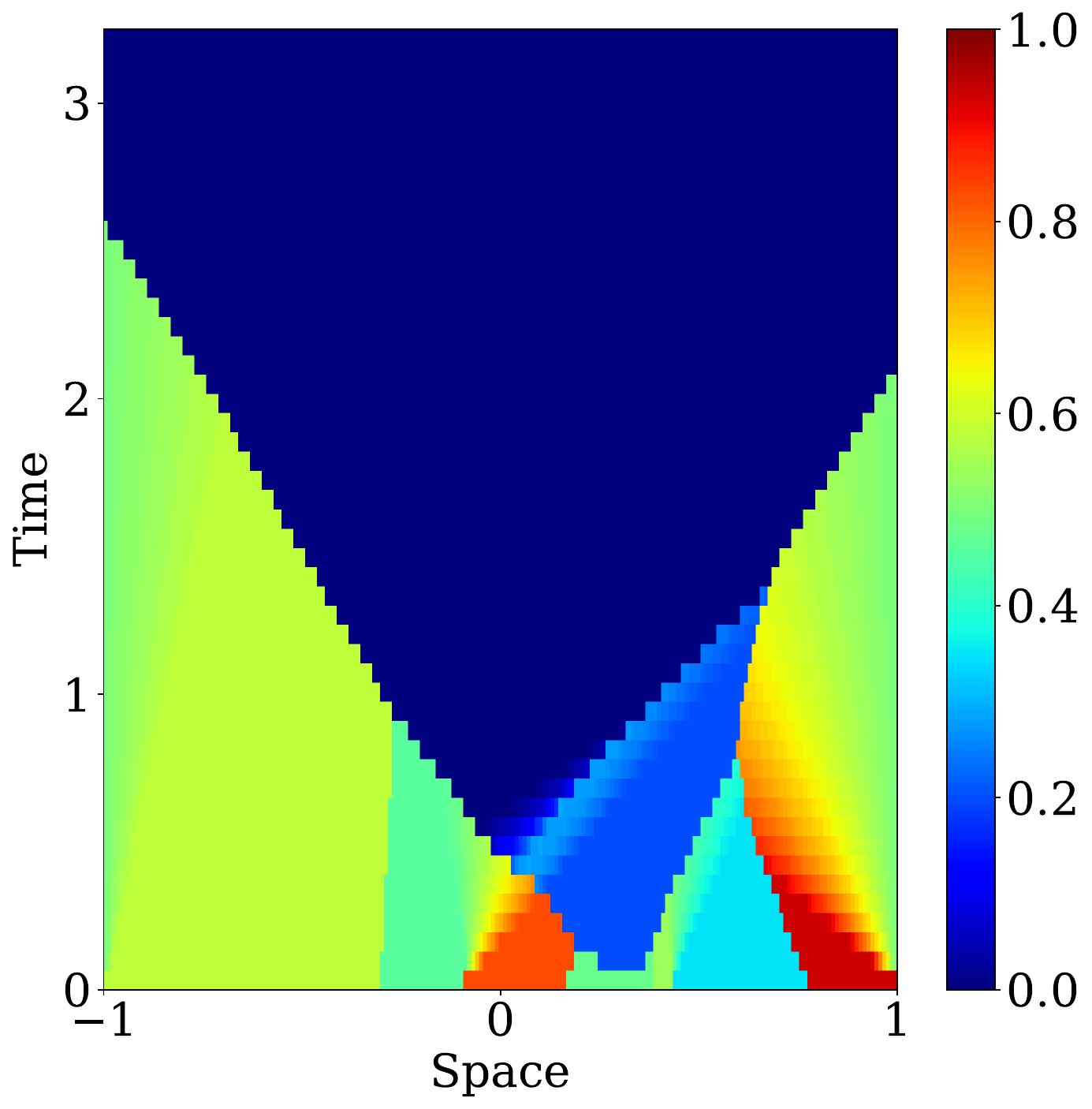}
    \caption{Real Data}
    \label{fig:real-data-wft-fno-sec7.1}
  \end{subfigure}
  \hspace{0.01\textwidth}
  \begin{subfigure}[b]{0.485\textwidth}
    \centering
    \includegraphics[width=\textwidth,height=0.31\textheight,keepaspectratio]{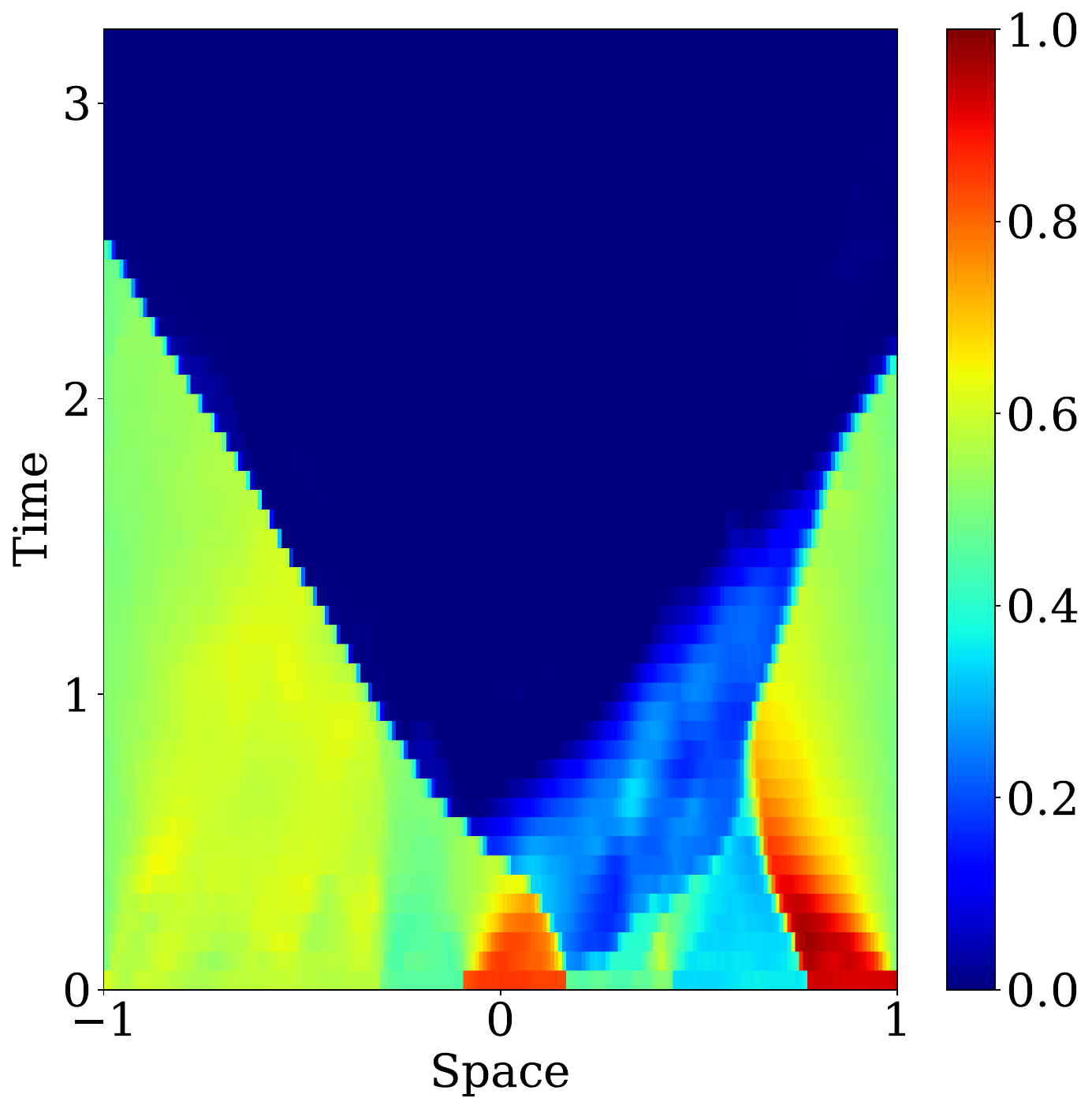}
    \caption{Predicted Data}
    \label{fig:predicted-data-wft-fno-sec7.1}
  \end{subfigure}
  \hspace{0.02\textwidth}
  \caption{Prediction results for data with six discontinuities in initial condition, generated using WFT algorithm and trained on FNO }
  \label{fig:wft-fno-sect7.1}
\end{figure}

Tables \ref{tab:wft l2 loss} and \ref{tab:l2 loss gd} present the relative \(L_2\) loss for the experiments carried out on the IVP using the data generated by the WFT algorithm and the Godunov scheme, respectively. In particular, the Godunov scheme exhibits a superior learning capacity compared to the WFT algorithm. For comparison, Figures \ref{fig:gd-fno-sect7.1} and \ref{fig:wft-fno-sect7.1} show the sample for the same initial condition. But, the relative $L_2$ loss for the given sample using Godunov scheme is $0.173$ while for the WFT algorithm it is $0.215$. This shows how FNO performs very poorly irrespective of the training data.

\begin{table}[htb]
\centering
\caption{Relative $L_2$ loss for data generated using WFT Algorithm $(\Delta \rho=1/250)$}
\label{tab:wft l2 loss}
\begin{tabular}{l|cc|cc|cc|cc}
\toprule
Data & \multicolumn{2}{c|}{Num\_Sample} & \multicolumn{2}{c|}{FNO} & \multicolumn{2}{c|}{MWT} & \multicolumn{2}{c}{WNO} \\
\cmidrule(lr){2-3} \cmidrule(lr){4-5} \cmidrule(lr){6-7} \cmidrule(lr){8-9}
     & \textit{Easy} & \textit{complex} & \textit{Easy} & \textit{complex} & \textit{Easy} & \textit{complex} & \textit{Easy} & \textit{complex} \\
\midrule
Train &  2960  & 8040   & 0.038 &  0.094     & 0.047 &  0.091     & 0.060 &    0.109   \\
Val   &  880  & 2400  & 0.078 &   0.126    & 0.085 &    0.127   & 0.076 &  0.126     \\
Test  & 880 & 2400 & 0.078 & 0.127 & 0.085 & 0.128 & $\mathbf{0.076}$ & $\mathbf{0.126}$ \\
\midrule
Time per Epoch (s) &    &   & $\mathbf{6.91}$ &  $\mathbf{18.78}$     & 13.54 &  29.48     & 42.98 &    111.34   \\
\bottomrule
\end{tabular}
\end{table}

\begin{table}[htb]
\centering
\caption{Relative $L_2$  loss for data generated using Godunov Scheme}
\label{tab:l2 loss gd}
\begin{tabular}{l|cc|cc|cc|cc}
\toprule
Data & \multicolumn{2}{c|}{Num\_Sample} & \multicolumn{2}{c|}{FNO} & \multicolumn{2}{c|}{MWT} & \multicolumn{2}{c}{WNO} \\
\cmidrule(lr){2-3} \cmidrule(lr){4-5} \cmidrule(lr){6-7} \cmidrule(lr){8-9}
      & \textit{Easy} & \textit{complex} & \textit{Easy} & \textit{complex} & \textit{Easy} & \textit{complex} & \textit{Easy} & \textit{complex} \\
\midrule
Train &  3000  & 8040   & 0.022 &  0.081     & 0.034 &   0.088    & 0.023 &   0.097   \\
Val   &  900  & 2400  & 0.064 &   0.113    & 0.071 &   0.118   & 0.064 &   0.113   \\
Test  &  900  & 2400  & 0.064 &   0.114    & 0.070 &     0.119  & $\mathbf{0.063}$ &   $\mathbf{0.113}$    \\
\midrule
Time per Epoch (s) &    &   & $\mathbf{5.04}$ &  $\mathbf{14.56}$     & 10.98 &   35.47    & 42.46 &  119.45    \\
\bottomrule
\end{tabular}
\end{table}

\begin{figure}[htbp]
  \centering
  \begin{subfigure}[b]{0.31\textwidth}
    \centering
    \includegraphics[width=\textwidth,height=0.31\textheight,keepaspectratio]{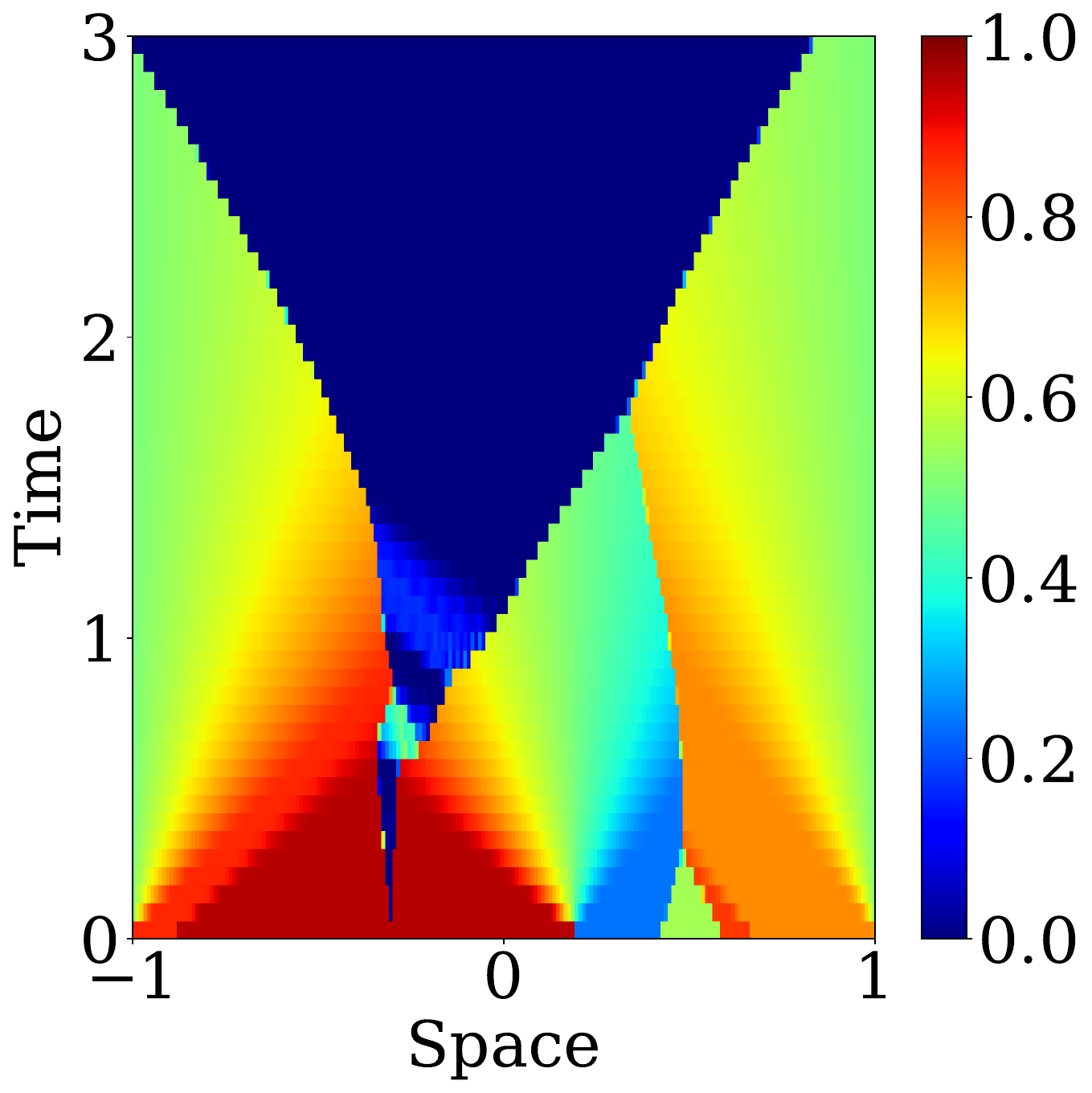}
    \caption{Godunov Scheme}
    \label{fig:real-data-gd-fno}
  \end{subfigure}
  \hspace{0.02\textwidth}
  \begin{subfigure}[b]{0.31\textwidth}
    \centering
    \includegraphics[width=\textwidth,height=0.31\textheight,keepaspectratio]{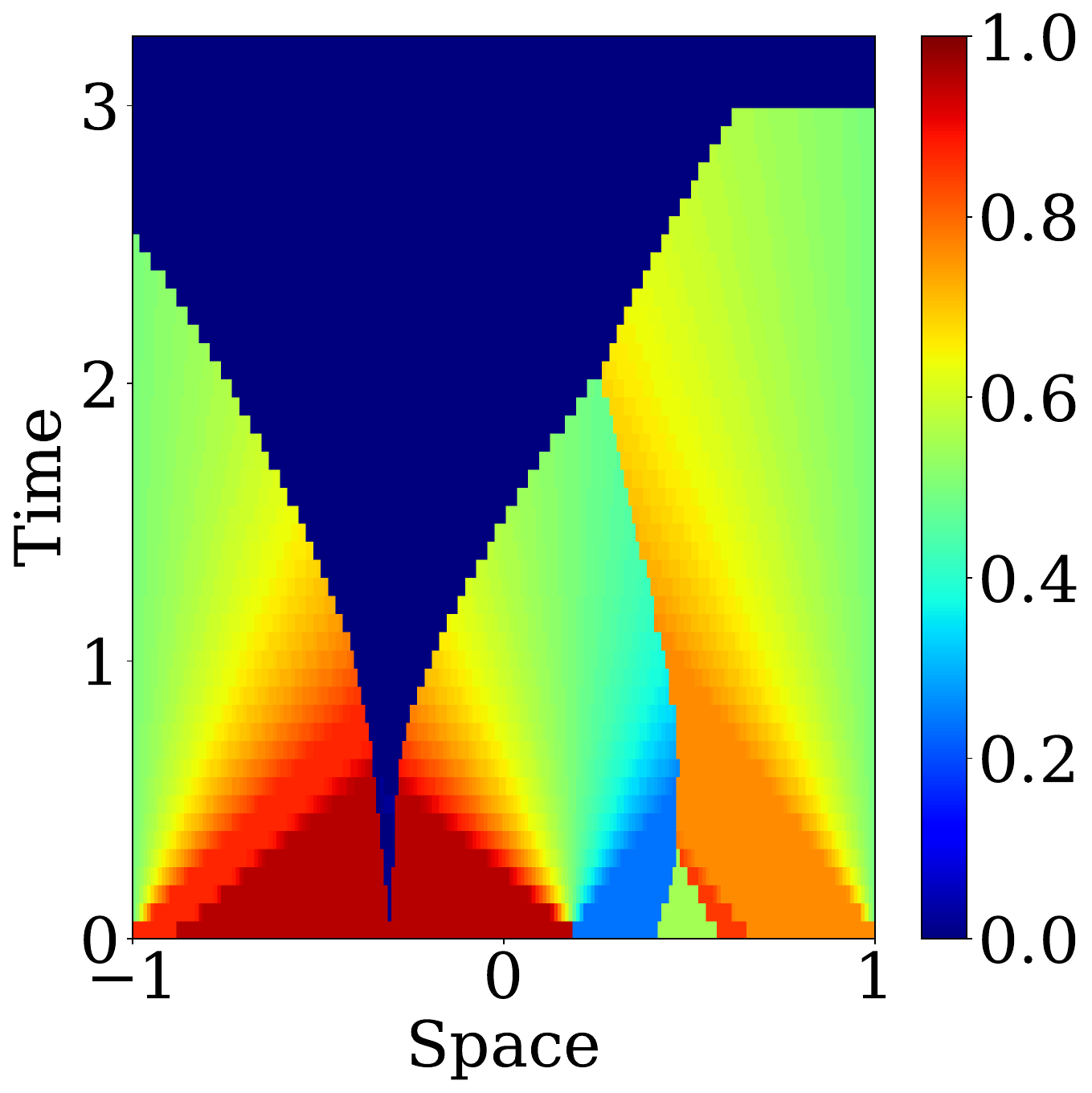}
    \caption{WFT with \(\Delta \rho = 1/250\)}
    \label{fig:real-data-wft_fno_250}
  \end{subfigure}
  \hspace{0.02\textwidth}
  \begin{subfigure}[b]{0.31\textwidth}
    \centering
    \includegraphics[width=\textwidth,height=0.31\textheight,keepaspectratio]{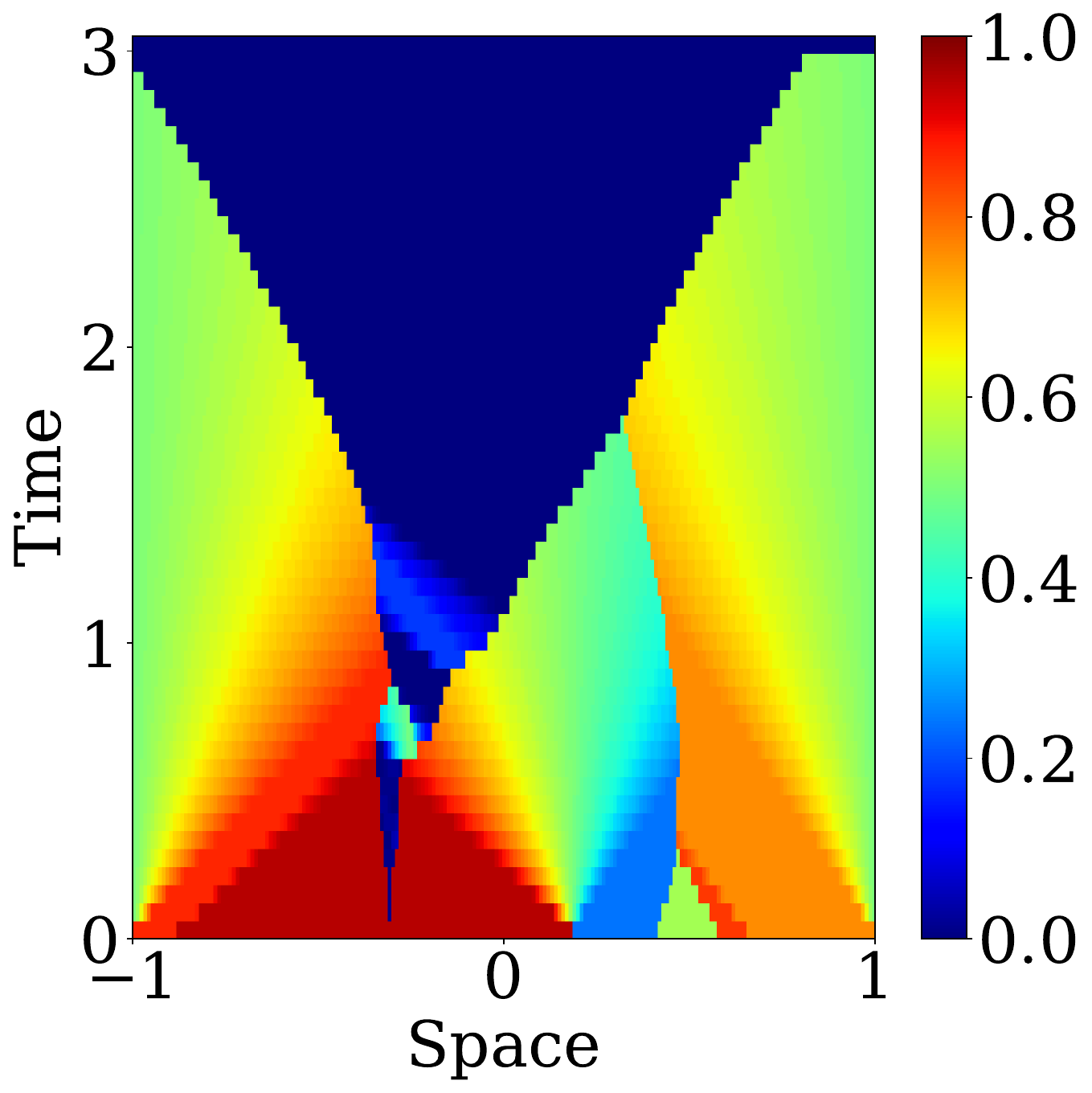}
    \caption{WFT with \(\Delta \rho = 1/800\)}
    \label{fig:real-data-wft_fno_800}
  \end{subfigure}
  \caption{Sample comparison for Goudnov and WFT algorithm}
  \label{fig:sample compariosn wft and gd}
\end{figure}

It is possible that the mesh size $\Delta \rho$ plays a key role in the observed failure of the learning process for the WFT algorithm. The mesh size \(\Delta \rho = 1/250\) used to generate the data used in Table \ref{tab:wft l2 loss} causes numerical instability for some samples.  
This numerical instability can be quantified by the "cost integral balance error" as follows.
\begin{equation}
    E_{\text{cost}} = \max_{t} \left| \int_{-1}^{\xi(t)} \frac{1}{1 - \rho(t, y)} \, dy - \int_{\xi(t)}^{1} \frac{1}{1 - \rho(t, y)} \, dy \right|.
\end{equation}
For example, Figure \ref{fig:sample compariosn wft and gd} shows the same sample generated by the Godunov and WFT scheme. Specifically, the sample depicted in Figure \ref{fig:real-data-wft_fno_250}, generated with \(\Delta \rho = 1/250\), exhibits numerical instability with $E_{\text{cost}}$ tending to infinity, consequently producing an unrealistic solution.. This shows the inconsistency in the data produced by the WFT algorithm, as compared to the Godunov scheme, leading to high relative $L_2$ loss.\\

Therefore, additional experiments were conducted using data generated by the WFT algorithm with \(\Delta \rho= 1/800\) and with samples having $E_{\text{cost}} < 0.1$. Figure~\ref{fig:real-data-wft_fno_800} illustrates that this approach yields a more accurate solution, which can potentially reduce the relative \(L_2\) loss. The results of this experiment are presented in Figure \ref{fig:fno_250,800}, where the relative $L_2$ loss is plotted against the number of discontinuities in the initial condition. It shows that even after significantly increasing the discretizations, the learning does not improve much. 
Hence, it can be concluded from Figure~\ref{fig:fno_250,800} that, as the number of discontinuities in the initial condition increases, the accuracy of neural operator declines significantly,reducing their effectiveness in complex scenarios. This shortfall suggests the need for alternative architectures that are better equipped to capture the discontinuities inherent in the model.
  
\begin{figure}[htbp]
  \centering
  \begin{subfigure}[b]{0.485\textwidth}
    \centering
    \includegraphics[width=\textwidth,height=0.31\textheight,keepaspectratio]{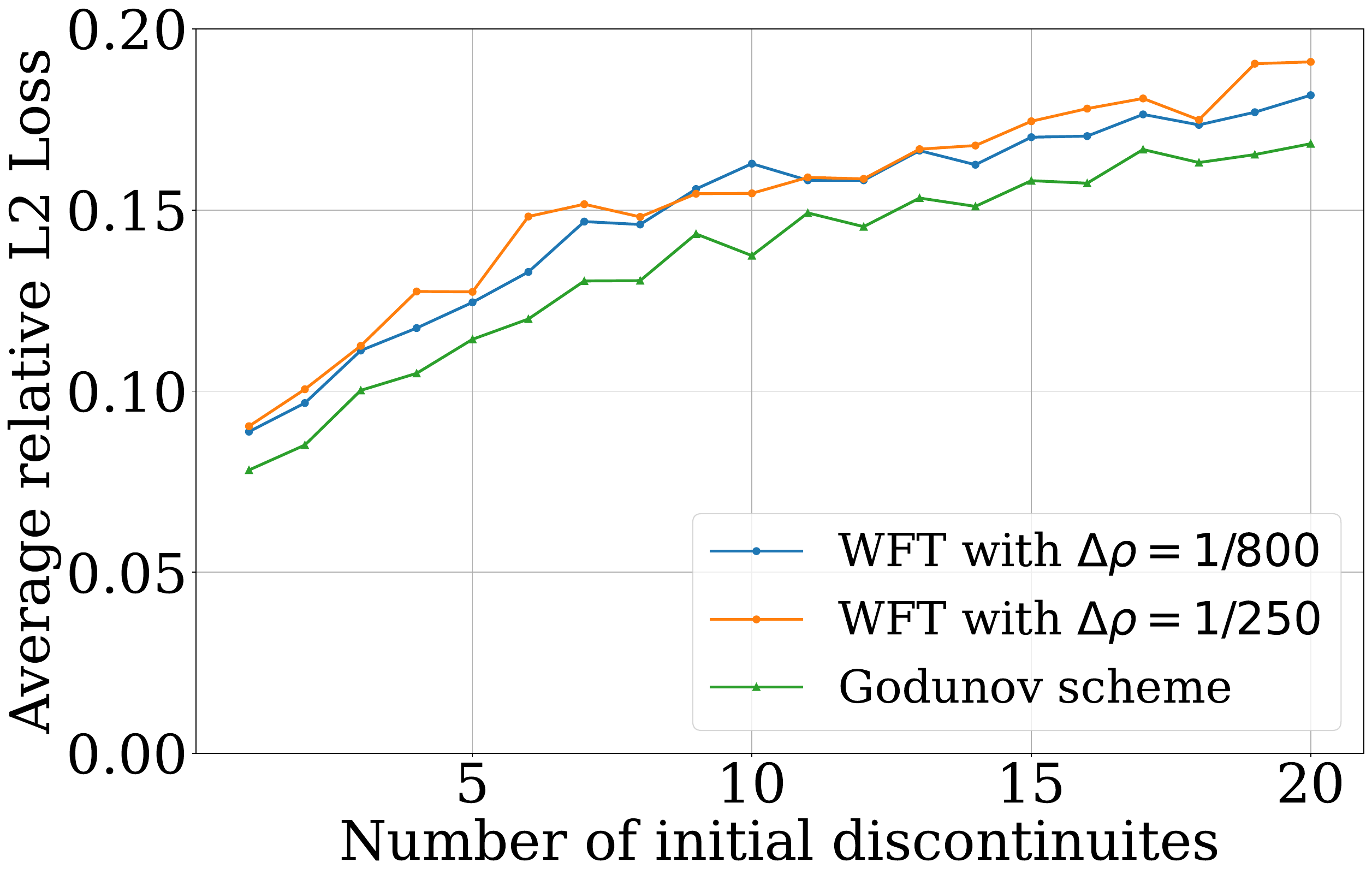}
    \caption{The evolution of relative $L_2$ loss w.r.t. initial discontinuities}
    \label{fig:fno_250,800}
  \end{subfigure}
  \hspace{0.01\textwidth}
  \begin{subfigure}[b]{0.485\textwidth}
    \centering
    \includegraphics[width=\textwidth,height=0.4\textheight,keepaspectratio]{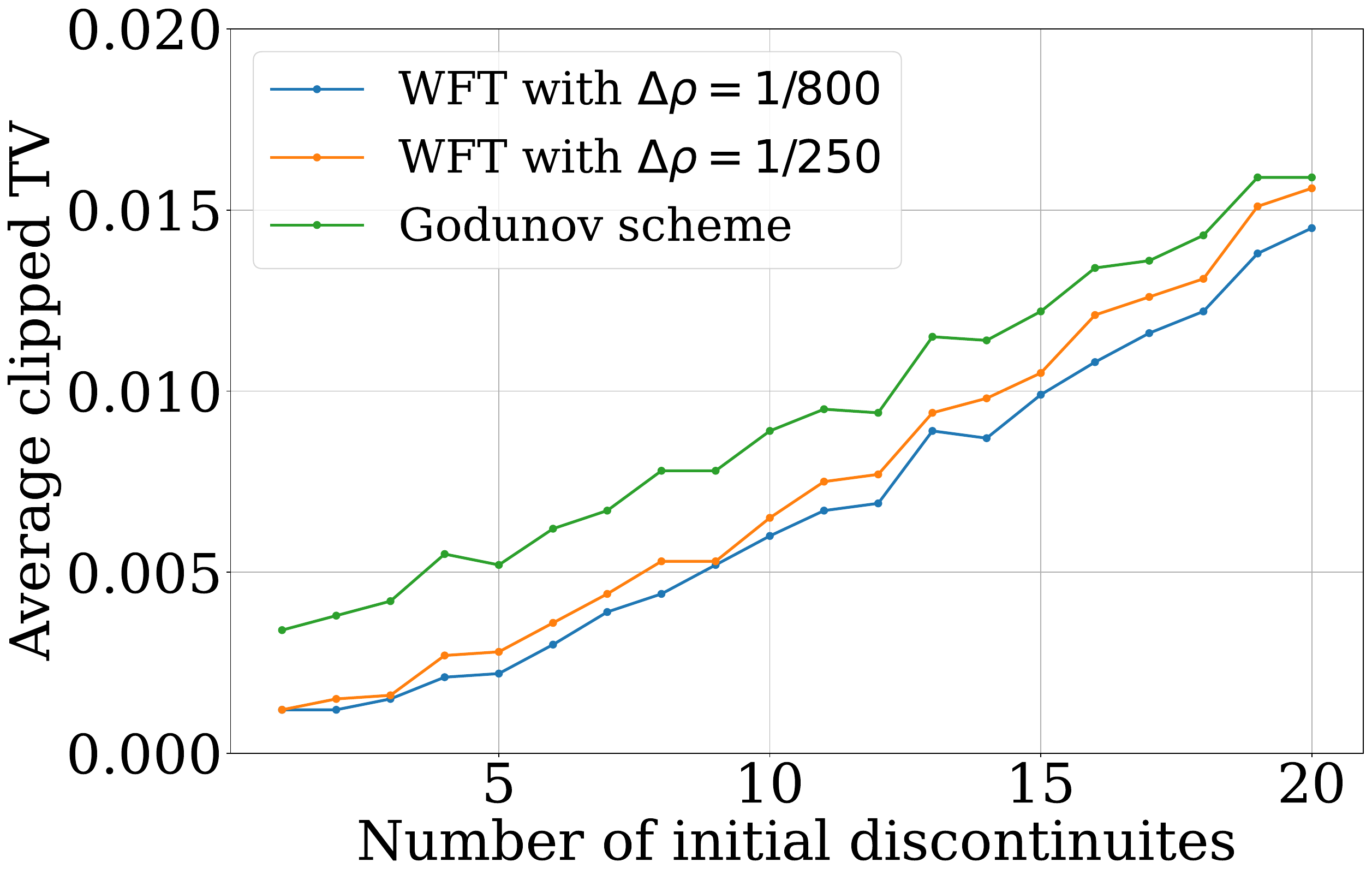}
   \caption{The evolution of clipped TV w.r.t. initial discontinuities}
    \label{fig:diff_error}
  \end{subfigure}
  \hspace{0.02\textwidth}
  \caption{Relative $L_2$ loss and diffusion error for \textit{complex} data generated by WFT and Godunov scheme, trained up to 10 discontinuities and tested upto 20 discontinuities in initial condition on FNO}
  \label{fig:l2 loss and diff error}
\end{figure}

\subsubsection{Problem II} \label{problem2}
This setup imposes time-invariant zero-flux boundary conditions, i.e., no mass enters or leaves the domain. This is achieved by assigning a very high value to the potential function as shown in equation (\ref{eq:bc_2}), at the domain boundaries, thereby effectively preventing agents from moving towards the exits. This provides insight into the operator’s ability to handle an accumulating domain. Training data is generated using the Godunov scheme for a scenario up to three discontinuities in the initial density profile. Consequently, the primary challenge for the operator is to capture and adapt to the effect of a fully closed boundary across time, rather than learning multiple boundary states or abrupt transitions. As a result, this scenario is expected to be relatively straightforward for the operator to learn, while still offering a meaningful test of its capability to handle non-trivial boundary conditions consistently. As this problem is still an initial value problem with closed boundary, the operator training data is structured as shown in Figure \ref{fig:training_data}, with different initial conditions.

\begin{figure}[h]
  \centering
  \begin{subfigure}[b]{0.31\textwidth}
    \centering
    \includegraphics[width=\textwidth,height=0.31\textheight,keepaspectratio]{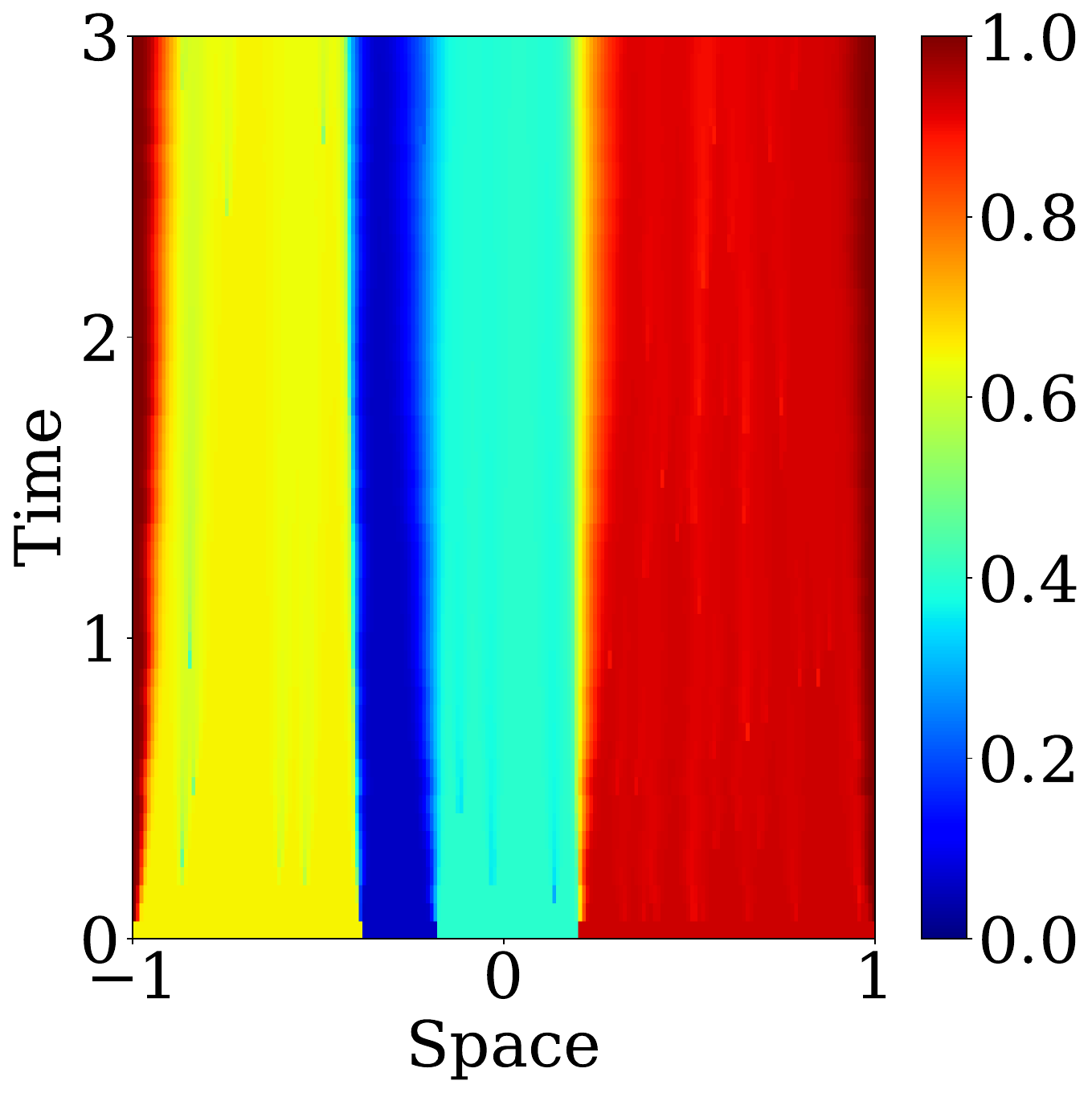}
    \caption{Real Data}
    \label{fig:real-data_bothclosed}
  \end{subfigure}
  \hspace{0.02\textwidth}
  \begin{subfigure}[b]{0.31\textwidth}
    \centering
    \includegraphics[width=\textwidth,height=0.31\textheight,keepaspectratio]{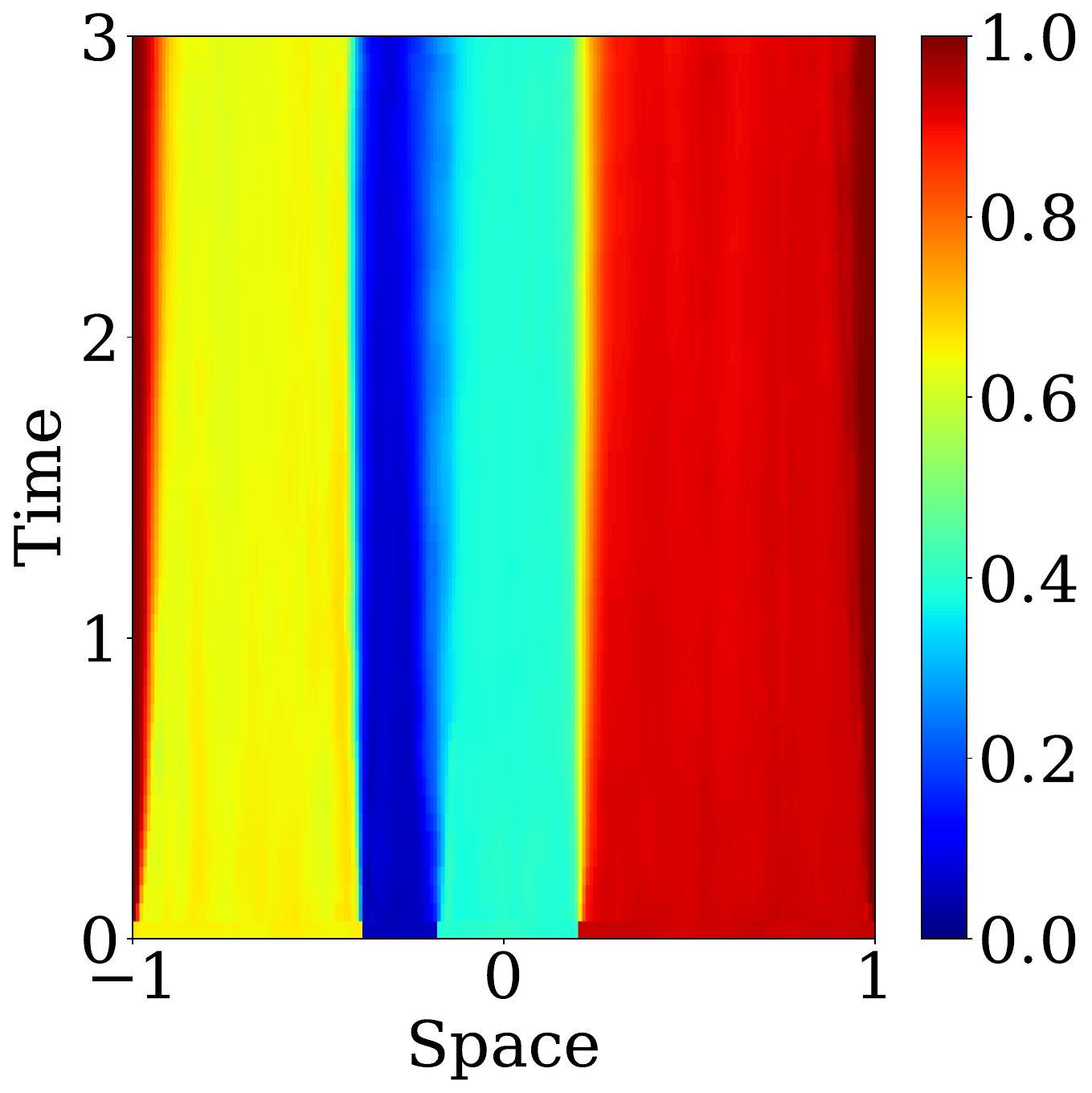}
    \caption{Predicted Data}
    \label{fig:predicted-data_bothclosed}
  \end{subfigure}
  \hspace{0.02\textwidth}
  \begin{subfigure}[b]{0.31\textwidth}
    \centering
    \includegraphics[width=\textwidth,height=0.31\textheight,keepaspectratio]{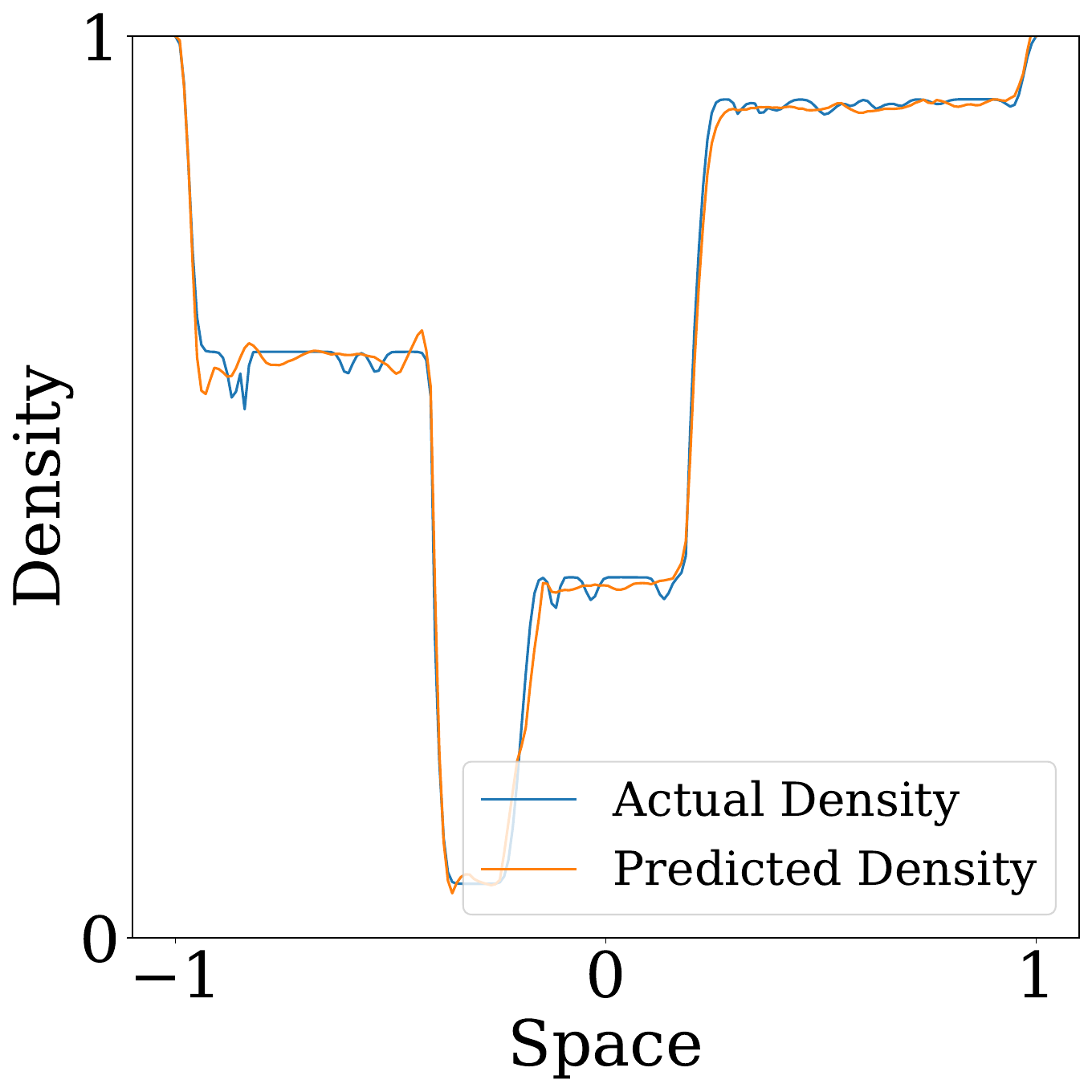}
    \caption{Predicted Density at $t=0.5$}
    \label{fig:predicted-density_bothclosed}
  \end{subfigure}
  \caption{Prediction results with three discontinuities in initial condition and boundary conditions given by equation \ref{eq:bc_2} (Train Loss: 0.032, Val Loss: 0.042, Test Loss: 0.042)}
  \label{fig:bothclosed_trained}
\end{figure}

Figure \ref{fig:bothclosed_trained} presents the results for the scenario in which both exits are closed. In this configuration, the boundary conditions are relatively straightforward, resulting in minimal boundary influence on the interior of the domain. Consequently, the learning performance is significantly high.

\subsubsection{Problem III} \label{problem3}
To isolate and analyze the impact of boundary conditions on the learning behavior of neural operators,  smooth, easy-to-learn initial conditions are used. Specifically, a zero discontinuity initial condition is constructed using a single Gaussian pulse with no discontinuities (see Subsection \ref{data_generate} for details). This ensures that the learning complexity is dominated by the boundary behavior rather than the initial condition itself. For this experiment, the dataset is generated using the Godunov scheme.

Given $\rho(t=0,x)$,  two distinct boundary scenarios are considered:
\begin{itemize}
    \item \textbf{Absorbing boundary condition on both exits:} Boundary condition given by equation (\ref{eq:bc_1})  were adopted. This allows density to flow freely out of
    the domain which helps in modeling the movement of pedestrians that exit the corridor.
    \item \textbf{Time varying absorbing boundary condition at left exit:} To simulate realistic and time-varying boundary control, boundary conditions were adopted as given by equation (\ref{eq:bc_3}).
\end{itemize}

These boundary conditions introduce time varying dynamics, enabling the evaluation of the neural operator's sensitivity to temporal variations in boundary condition. Using a smooth  initial condition (e.g. a Gaussian profile with no discontinuities), the complexity of the learning task is primarily driven by the boundary dynamics rather than the initial condition itself. For training purposes, each sample is structured as illustrated in Figure \ref{fig:BC II dataset}. \\

\begin{figure}[htbp]
  \centering
  \begin{subfigure}[b]{0.31\textwidth}
    \centering
    \includegraphics[width=\textwidth,height=0.31\textheight,keepaspectratio]{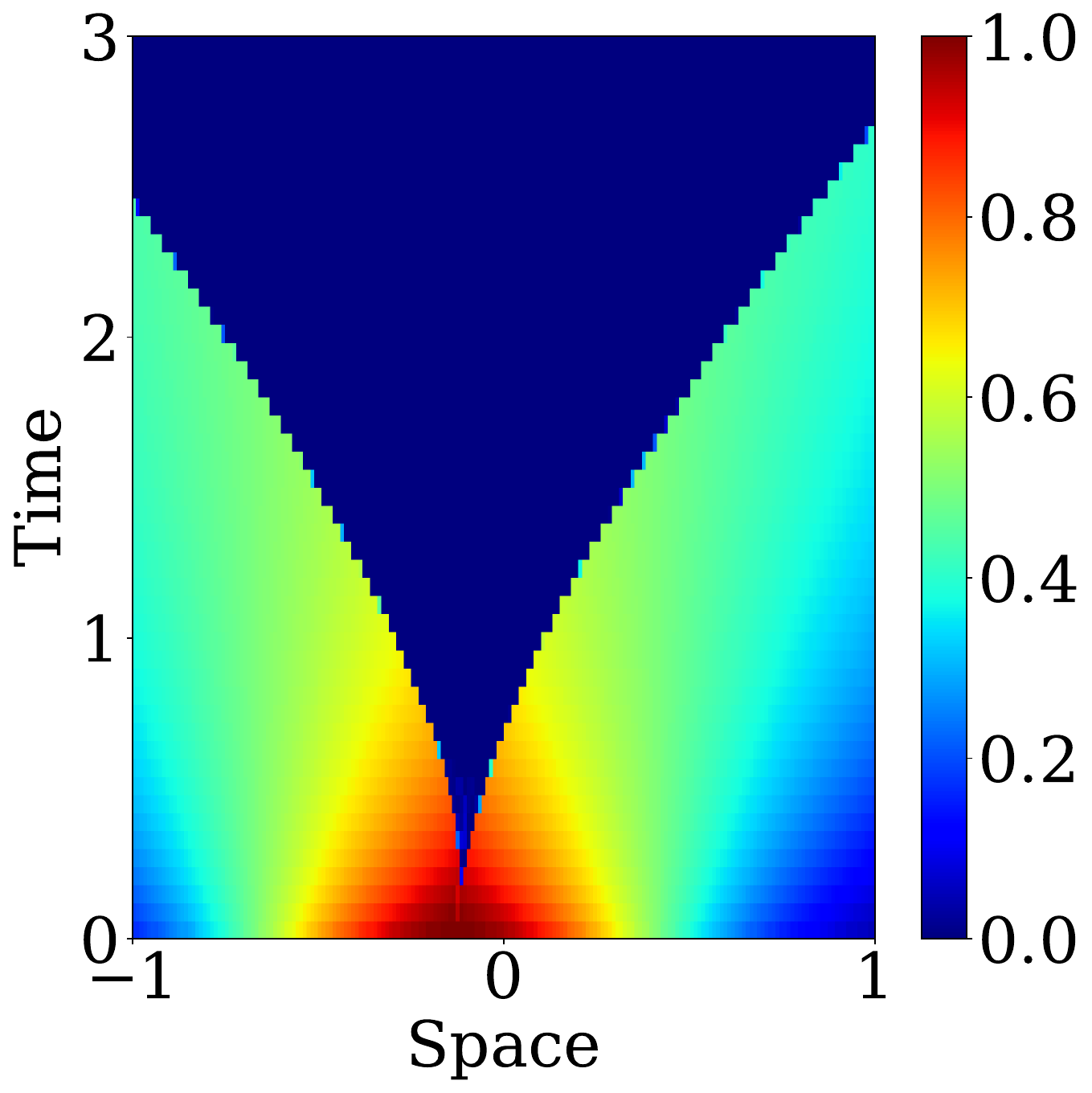}
    \caption{Real Data}
    \label{fig:real-data_gaussian}
  \end{subfigure}
  \hspace{0.02\textwidth}
  \begin{subfigure}[b]{0.31\textwidth}
    \centering
    \includegraphics[width=\textwidth,height=0.31\textheight,keepaspectratio]{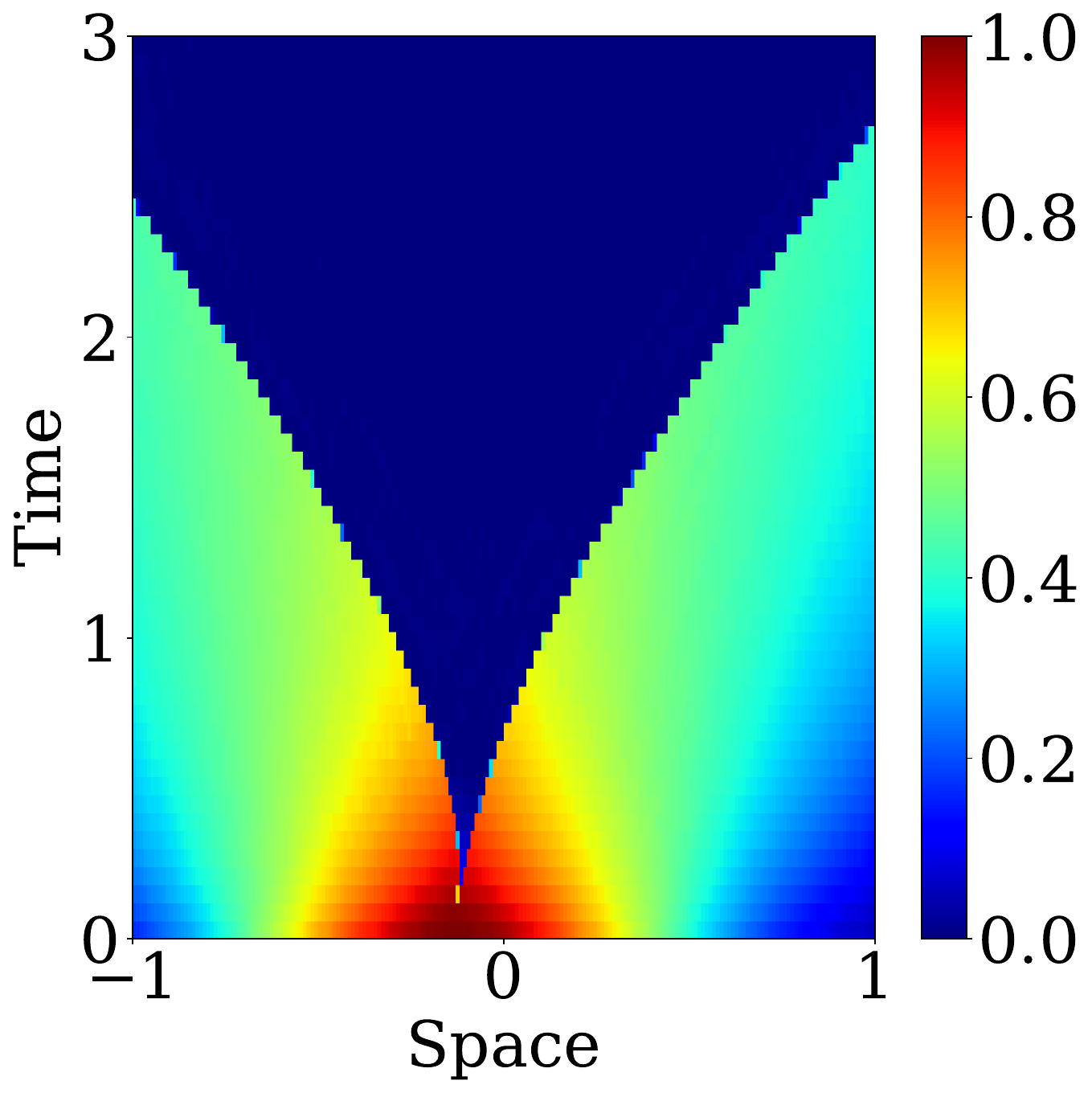}
    \caption{Predicted Data}
    \label{fig:predicted-data_gaussian}
  \end{subfigure}
  \hspace{0.02\textwidth}
  \begin{subfigure}[b]{0.31\textwidth}
    \centering
    \includegraphics[width=\textwidth,height=0.31\textheight,keepaspectratio]{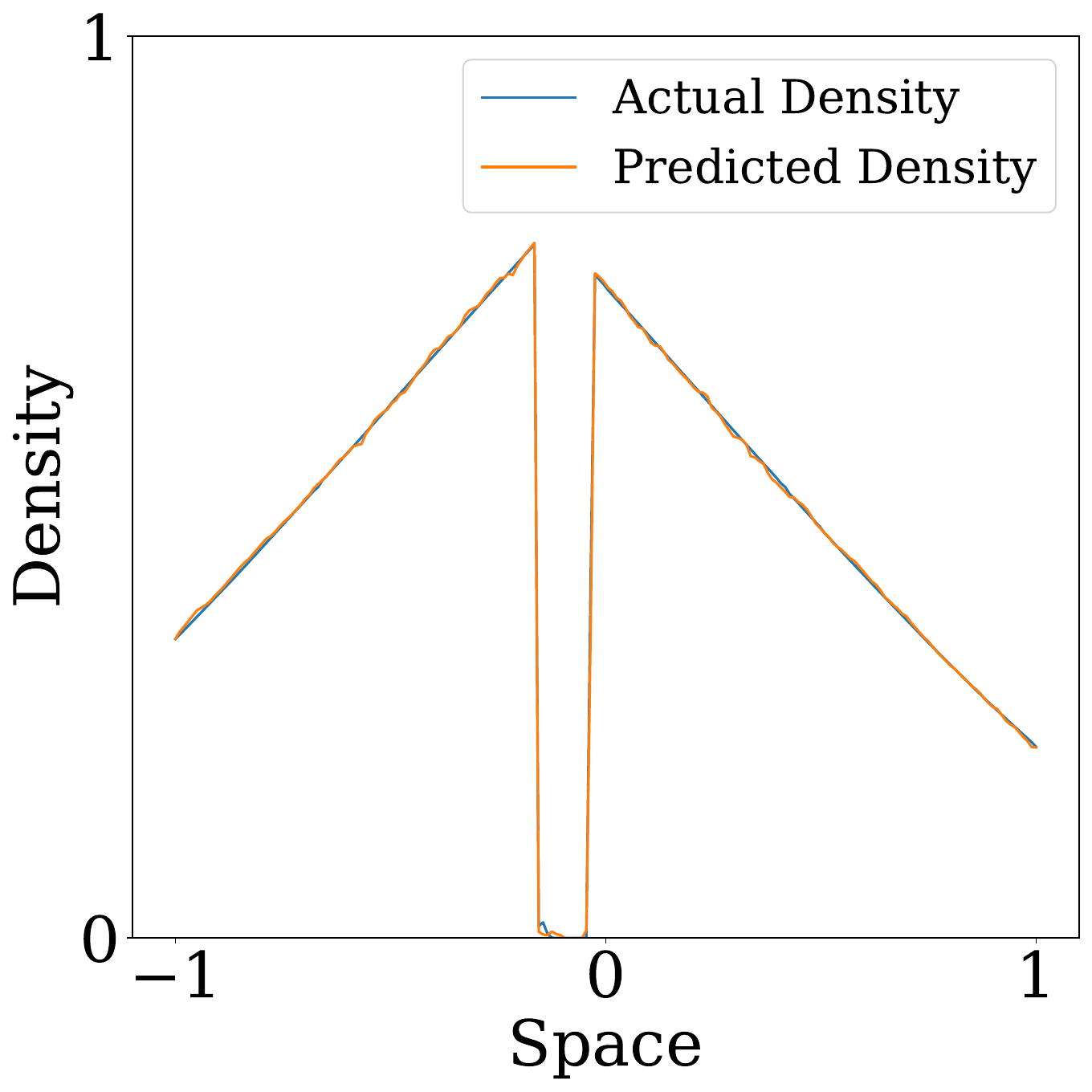}
    \caption{Predicted Density at $t=0.5$}
    \label{fig:predicted-density_gaussian}
  \end{subfigure}
  \caption{Prediction results with Gaussian initial condition}
  \label{fig:initial-gaussian}
\end{figure}

\begin{figure}[htbp]
  \centering
  \begin{subfigure}[b]{0.31\textwidth}
    \centering
    \includegraphics[width=\textwidth,height=0.31\textheight,keepaspectratio]{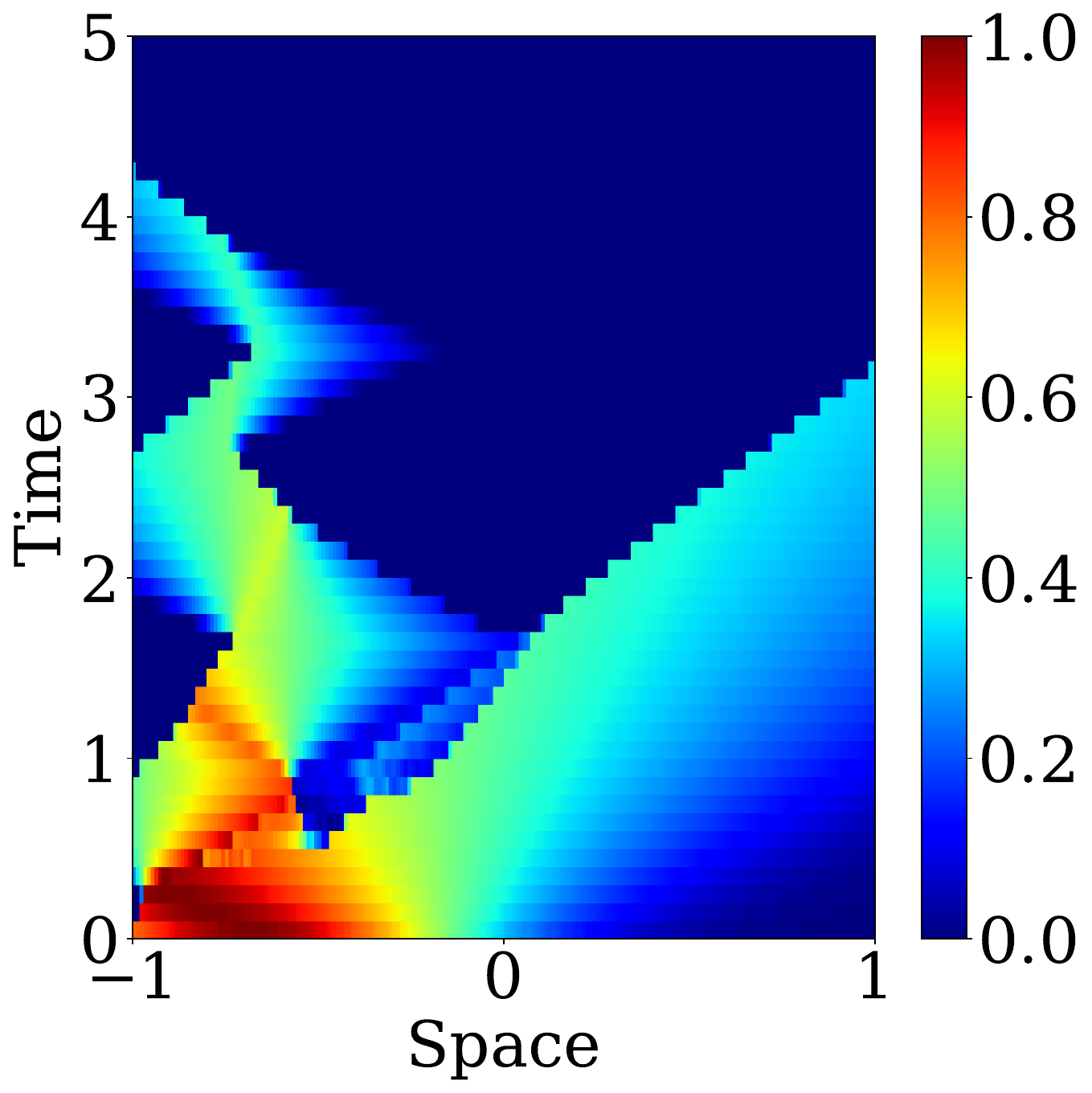}
    \caption{Real Data}
    \label{fig:real-data_closedBC}
  \end{subfigure}
  \hspace{0.02\textwidth}
  \begin{subfigure}[b]{0.31\textwidth}
    \centering
    \includegraphics[width=\textwidth,height=0.31\textheight,keepaspectratio]{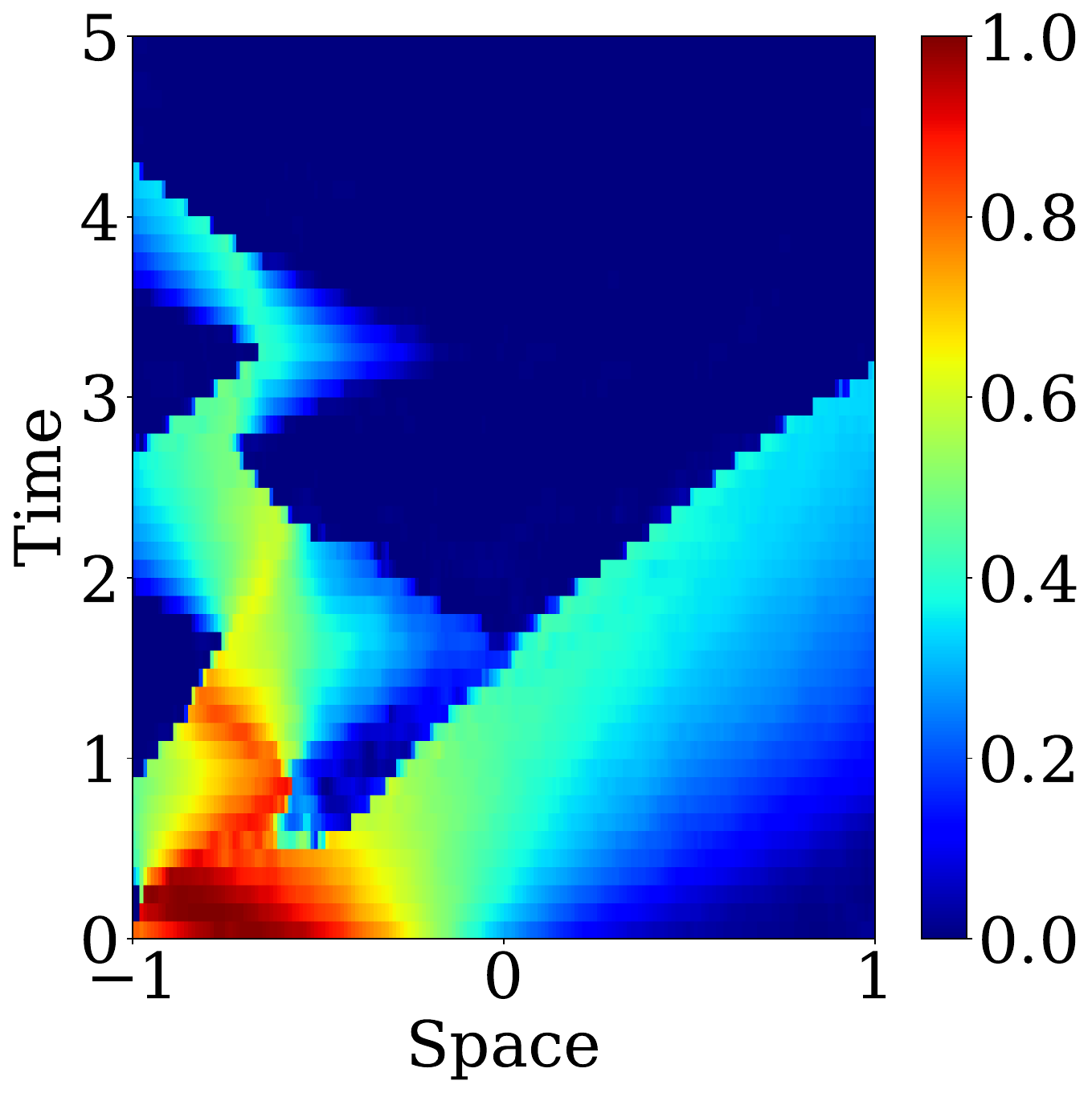}
    \caption{Predicted Data}
    \label{fig:predicted-data_closedBC}
  \end{subfigure}
  \hspace{0.02\textwidth}
  \begin{subfigure}[b]{0.31\textwidth}
    \centering
    \includegraphics[width=\textwidth,height=0.31\textheight,keepaspectratio]{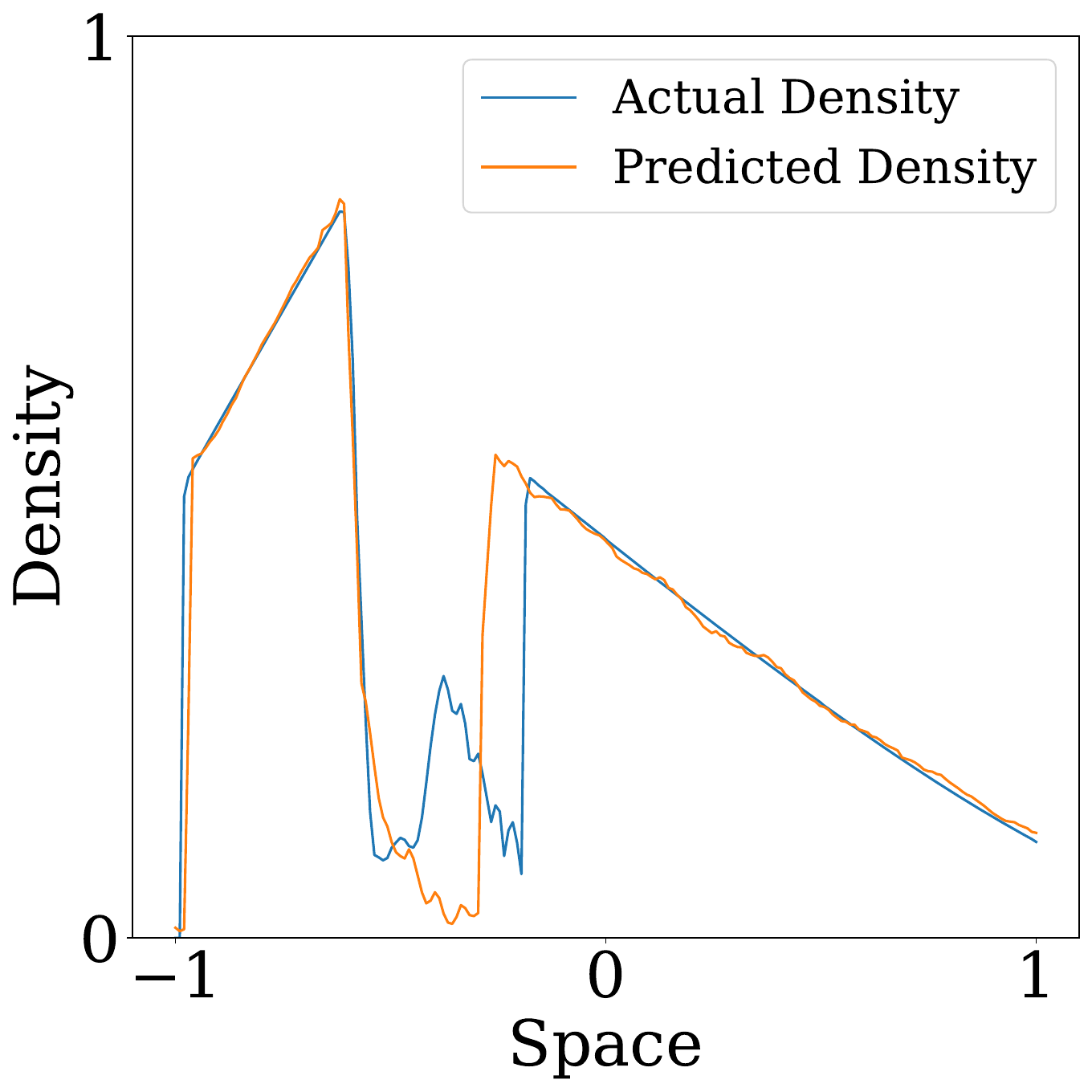}
    \caption{Predicted Density at $t=0.9$}
    \label{fig:predicted-density_closedBC}
  \end{subfigure}
  \caption{Prediction results under Gaussian initial conditions and boundary conditions given by equation \ref{eq:bc_3}.}
  \label{fig:initial-gaussian_closedBC}
\end{figure}

\begin{figure}
    \centering
    \includegraphics[width=\linewidth, height=0.4\textheight, keepaspectratio]{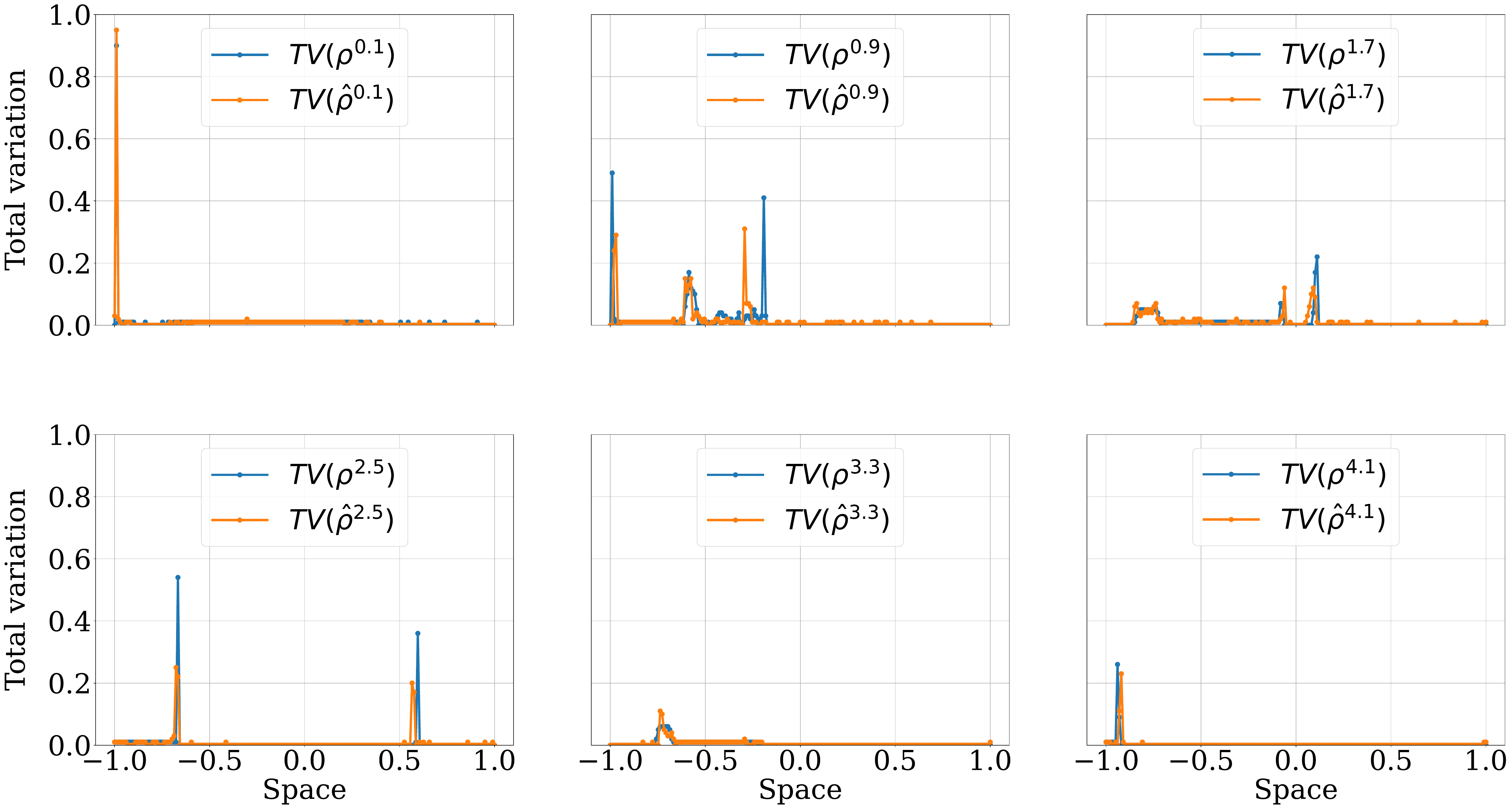}
    \caption{Total variation for the example sample shown in Figure \ref{fig:initial-gaussian_closedBC} at different time steps trained on FNO}
    \label{fig:tv_plots}
\end{figure}

Table \ref{tab:gaussian_table} shows the results for the case with Gaussian initial condition, where two cases were considered. Figure \ref{fig:initial-gaussian} shows the prediction results for the case when both exits are open. Figure \ref{fig:initial-gaussian_closedBC}, shows the results for the case where the left exit is closed non-periodically. For this case, a longer time interval is used to ensure that the density inside the domain goes to zero before simulation ends. This is because, if density remains non-zero inside the domain, any turning behavior exhibited will not be captured by the boundary condition unless it actually reaches the boundary.
For time varying absorbing boundary condition, two configurations were used as input to train the operators:
\begin{itemize}
    \item[\textbf{(i)}] Initial Value Problem (IVP): using only the initial density profile as training input.
    \item[\textbf{(ii)}] Mixed Initial-Boundary Value Problem (MI-BVP): Combining the initial density profile with the boundary conditions as training input.
\end{itemize}
The results for these experiments are presented in Table~\ref{tab:gaussian_table} (columns 4 and 5, respectively). Here we see that the boundary conditions challenge the kernel-based representation of the neural operator by violating its core assumptions. The solution operator is no longer a simple translation-invariant convolution on the domain; it includes boundary integral terms and position dependent effects. A neural operator ignorant of this will tend to produce poor approximations, especially in scenarios where the solution is dominated by boundary driven effects.
When boundary effects dominate, the shortcomings of the Green’s function assumption become more pronounced. The presence of boundaries means the true solution operator often decomposes into multiple components rather than a single integral kernel. Neural operators often approximate PDE solutions via a single kernel integral
where  the kernel acts like a Green's function learned from data. In problems with
nearly fixed or smoothly varying boundary conditions, a single integral kernel
can capture the solution's global behavior. However, when boundary conditions
switch abruptly (e.g., an exit that randomly opens or closes), the actual PDE solution
exhibits distinct regimes that a single static kernel struggles to learn. Relying on a single integral kernel
forces the model to ``average'' conflicting boundary dynamics, often resulting
in poor generalization as can be seen in Figure \ref{fig:initial-gaussian_closedBC}. As further illustrated in Figure \ref{fig:tv_plots}, the NO approximation treats discontinuities inconsistently, dampening some jumps, exaggerating others, and occasionally missing them altogether, reinforcing the earlier assertion of its difficulty in capturing sharp features accurately.

\begin{table}[ht]
  \centering
  \caption{Relative $L_2$  loss for Data with Gaussian Initial Condition}
  \label{tab:gaussian_table}
  \begin{tabular}{l | c | c|  c | c}
    \toprule
    Data & 
      \parbox{2cm}{\centering Num\_sample } & 
      \parbox{2.5cm}{\centering Both exit open} & 
      \parbox{3cm}{\centering Left exit closed non-periodically\\ (\textit{IVP})} & 
      \parbox{3cm}{\centering Left exit closed non-periodically\\ (\textit{MI-BVP})} \\
    \midrule
    Train & 1000 & 0.042 & 0.167 & 0.042 \\
    Val   & 300 & 0.060 & 0.232 & 0.155 \\
    Test  & 300  & 0.053 & 0.220 & 0.155 \\
    \midrule
    Time per Epoch (s) & & 1.90 & 1.96 & 1.96 \\
    \bottomrule
  \end{tabular}
\end{table}

\subsection{Numerical time-complexity}
In this section, we compare the computational complexity of various methods used to solve the Hughes model across different discretizations. From the NO class, the FNO method is analyzed as it provides the fastest computations while maintaining a favorable trade-off with accuracy. A forward pass is performed on the entire test set (2400 samples), and the average computation time per sample is calculated. For both the Godunov scheme and the WFT algorithm, 150 samples are generated, and the average time per sample is calculated. The results are as shown in Table \ref{tab:time_comparison}. Although the FNO outperforms traditional numerical schemes in terms of computational efficiency, its accuracy remains insufficient for reliably generating solutions to complex PDEs (Hughes model in this case). This gap motivates further exploration of neural operators that can achieve both high accuracy and efficiency.
\begin{table}[ht]
\centering
\begin{tabularx}{\textwidth}{l X c}
\toprule
\textbf{Method} & \textbf{Parameters} & \textbf{Average time to solve given initial condition (ms)} \\
\midrule
FNO & $\Delta x=1/200$, $\Delta t=1/50$  & \begin{tabular}[c]{@{}l@{}}Batch size = 1: $\mathbf{6}$\\ Batch size = 100: $\mathbf{0.4}$\end{tabular} \\
\midrule
Godunov & $\Delta x=1/100$, $\Delta t=1/100$    & 74  \\
& $\Delta x=1/400$, $\Delta t=1/400$    & 2161  \\
        & $\Delta x^*=1/500$, $\Delta t^*=1/600$    & 4740  \\
\midrule
WFT Algorithm & $\Delta \rho=1/100$  & 95 \\
& $\Delta \rho^*=1/250$  & 730 \\
              & $\mbox{} \quad \Delta \rho*=1/800$ & 2267 \\
\bottomrule
\end{tabularx}
\caption{Comparison of average time per sample and numerical parameters for different methods.\newline * Data at this discretization has been used for training.}
\label{tab:time_comparison}
\end{table}

\section{Conclusion}

In summary, our results reveal key limitations of current state-of-the-art neural operator when applied to complex nonlinear hyperbolic PDEs, such as the Hughes model for crowd dynamics. Although the tested neural operators perform reasonably well under simple conditions, their accuracy degrades noticeably as the complexity of initial and boundary conditions increases. This decline highlights a fundamental issue: existing architectures are not well-equipped to approximate solutions that contain multiple discontinuities, turning points, or nonlocal interactions, all of which are characteristic of the Hughes model. Moreover, we observe that the neural operators frequently fail to capture the precise location and strength of shock‑like discontinuities; in many cases they either under‑predict the sharpness or, conversely, artificially amplify or smooth out these abrupt transitions, leading to qualitatively incorrect flow patterns.

This study points to the need for new neural operator models or training strategies that are better suited to handling the challenges posed by nonlinear inviscid hyperbolic systems. Future research should focus on frameworks capable of handling abrupt transitions, thereby enhancing the robustness and reliability of predictive models in real-world applications.

\section{Acknowledgment}
This work was supported in part by the NYUAD Center for Interacting Urban Networks (CITIES), funded by Tamkeen under the NYUAD Research Institute Award CG001, and in part by the NYUAD Research Center on Stability, Instability, and Turbulence (SITE), funded by Tamkeen under the NYUAD Research Institute Award CG002. The views expressed in this article are those of the authors and do not reflect the opinions of CITIES, SITE, or their funding agencies.
\section{Appendix}

\subsection{Hyperparameters}

The hyperparameters for all the neural operators are provided in Table~\ref{tab:NOs hyperparams}.

\begin{table}[htb]
\centering
\caption{Hyperparameters for FNO, MWT, and WNO models.}
\label{tab:NOs hyperparams}
\begin{tabular}{lccc}
\hline
\textbf{Hyperparameter} & \textbf{FNO} & \textbf{MWT} & \textbf{WNO} \\
\hline
Batch size            & 40            & 40                    & 40 \\
Learning rate          & 0.01          & 0.01                  & 0.01 \\
Epochs                  & 500           & 500                   & 500 \\
Step size              & 100           & 100                   & 100  \\
Weight decay rate                  & 0.5           & 0.5                   & 0.5  \\
Fourier modes along t (modes\_t)                  & 10            & --                    & -- \\
Fourier modes along x (modes\_x)                 & 40            & --                    & -- \\
Projection dimension                  & 80            & --                    & 80 \\
Sample shape (t,x) WFT          & 50$\times$201       & 128$\times$128      & 49$\times$199 \\
Sample shape  (t,x) Godunov         & 50$\times$200       & 128$\times$128      & 49$\times$199 \\
Weight decay           & 1e-3          & 1e-3                  & 1e-3 \\
Grid range (t,x)            & [0:3,0:2]            & [0:3,0:2]                     & [0:3,0:2]  \\
\hline
\end{tabular}
\end{table}

\begin{figure}
    \centering
    \includegraphics[width=0.65\linewidth, height=0.35\textheight, keepaspectratio]{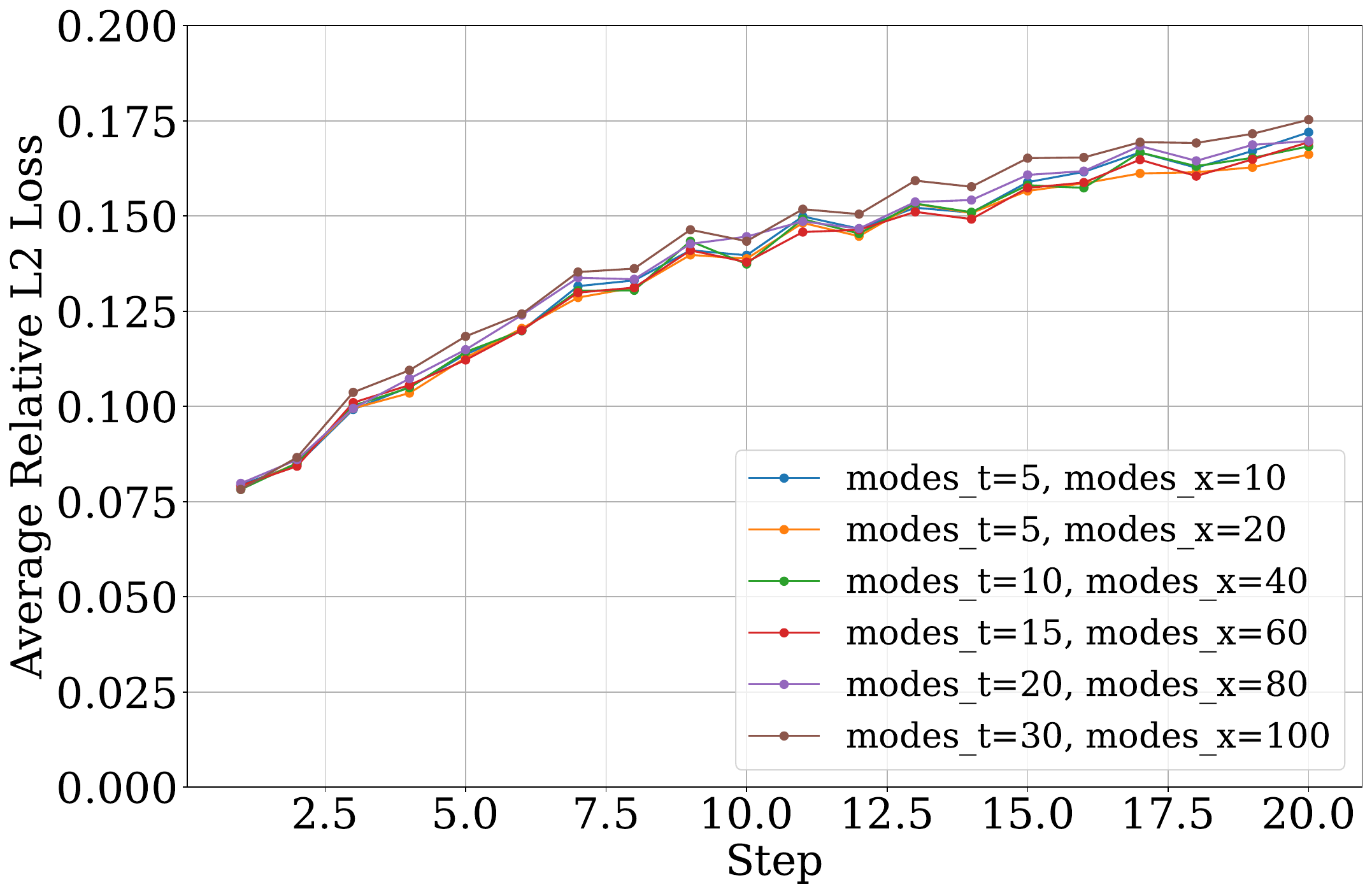}
    \caption{Relative $L_2$ loss for \textit{complex} data generated by Godunov scheme and trained up to ten discontinuities in initial condition on FNO with various Fourier modes}
    \label{fig:Fourier modes}
\end{figure}

\subsection{Algorithm: Godunov scheme for Hughes model}\label{sec:godunov algorithm}

\begin{algorithm}[H]
\caption{Data generation using Godunov scheme}
\begin{algorithmic}[1]
\State \textbf{Input:} Grid \(x=(x_1,x_2,\dots,x_n)\), Initial density \(\rho(t=0,x)\), Time step \(\Delta t\), Spatial step \(\Delta x\), Final time \(T\)
\State \textbf{Output:} Density \(\rho(x,t)\)
\State Initialize: \(t \gets 1\)
\While{\(t < T\)}
    \State \textbf{Compute turning point using cost integral balance},
    \For{each grid index \(x_i\)}
        \State \(C_{\text{left}}(x_i) \gets \Delta x \sum_{j=x_1}^{x_i} \frac{1}{1-\rho(t-1,x_j)} + \phi_{bc1}(t)\)  \quad \(C_{\text{right}}(x_i) \gets \Delta x \sum_{j=x_i}^{x_n} \frac{1}{1-\rho(t-1,x_j)} +\phi_{bc2}(t)\)
    \EndFor

    \State Find index \(x_{i^*}\) that minimizes \(|C_{\text{left}}(x_i) - C_{\text{right}}(x_i)|, \text{ note }, \phi_{bc1}(t)= \phi_{bc2}(t) =0\)
    \State Set turning point: \(\xi(t) \gets x_{i^*}\)
    \State \textbf{Define subdomains: left and right of \(\xi(t)\)}
    \State With equal probability, choose one of the following:\\
         \(\Omega_L = \{ x \le \xi(t) \}\) and \(\Omega_R = \{ x > \xi(t) \}\) or \(\Omega_L = \{ x < \xi(t) \}\) and \(\Omega_R = \{ x \geq \xi(t) \}\)
    \State \textbf{Update right region \(\Omega_R\) using Godunov scheme}
    \For{each cell \(x_i\) in \(\Omega_R\) with \(i = i^*, i^*+1, \dots, n\)}
        \State Compute numerical flux \(q_{i+1/2} = F(\rho(t-1,x_i),\rho(t-1,x_{i+1}))\)
        \State Update:
        \[
        \rho(t,x_i) \gets \rho(t-1,x_i) - \frac{\Delta t}{\Delta x}\Bigl(q_{i+1/2} - q_{i-1/2}\Bigr)
        \]
    \EndFor

    \State \textbf{Update left region \(\Omega_L\) using Godunov scheme}
    \For{each cell \(x_i\) in \(\Omega_L\) with \(i = i^*, i^*-1, \dots, 1\)}
        \State Compute numerical flux \(q_{i-1/2} = F(\rho(t-1,x_i),\rho(t-1,x_{i-1}))\)
        \State Update:
        \[
        \rho(t,x_i) \gets \rho(t-1,x_i) - \frac{\Delta t}{\Delta x}\Bigl(q_{i-1/2} - q_{i+1/2}\Bigr)
        \]
    \EndFor
    \State \(t \gets t + \Delta t\)
\EndWhile
\State \textbf{Return} \(\rho(x,t)\)
\end{algorithmic}
\end{algorithm}

\end{document}